\documentclass[letterpaper, 10 pt, conference]{ieeeconf}  
\usepackage{booktabs}
\usepackage{adjustbox}
\usepackage{graphicx}
\usepackage{amsmath}
\usepackage{amssymb}
\usepackage[abs]{overpic}
\usepackage{tabularx}
\usepackage{xcolor}
\usepackage{balance}
\usepackage{hyperref}

\IEEEoverridecommandlockouts                              

\title{\LARGE \bf
Visual Relocalization from Sparse Views in Aliased and Low-Texture Environments via Novel View Synthesis
}

\author{Maria Peribañez$^{1}$, Javier Civera$^{1}$, Rudolph Triebel$^{2,3}$ and Riccardo Giubilato$^{2}$
\thanks{This work was supported by the Helmholtz Association project iFOODis (contract number KA2-HSC-06), by the German Federal Ministry of Research, Technology and Space (BMFTR) under the Robotics Institute Germany (RIG), by Spanish projects PID2024-155886NB-I00 PID2024-158322OB-I00 (MCIN/ AEI/10.13039/ 501100011033, FEDER/UE and NextGenerationEU/PRTR) and by the Aragón Government via project T45\_23R.}
\thanks{$^{1}$Maria Peribañez, and Javier Civera are with University of Zaragoza, Spain
        {\tt\small \{mperibanez,jcivera\}@unizar.es}}%
\thanks{$^{2}$ Riccardo Giubilato and Rudolph Triebel are with German Aerospace Center (DLR), Institute of Robotics and Mechatronics
        {\tt\small \{Riccardo.Giubilato,rudolph.triebel\}@dlr.de}}%
\thanks{$^{3}$ Rudolph Triebel is also with Karlsruhe Institute of Technology (KIT), Karlsruhe, Germany
        {\tt\small rudolph.triebel@kit.de}}%
}

\begin{document}

\newcommand{\RG}[1]{\textcolor{red}{\textbf{#1}}}
\newcommand{\M}[1]{\textcolor{blue}{\textbf{#1}}}
\newcommand{\JC}[1]{\textcolor{green}{\textbf{#1}}}
\maketitle
\thispagestyle{empty}
\pagestyle{empty}

\begin{abstract}

Visual localization becomes extremely challenging in planetary-like terrains characterized by low texture, perceptual aliasing, harsh illumination, and sparse, weakly overlapping viewpoints induced by forward rover motion and unconstrained driving directions. Under these conditions, state-of-the-art image-to-image and image-to-map matching pipelines suffer significant performance degradation. In this work, we propose a visual relocalization method that departs from classical correspondence-based pipelines by directly estimating camera poses against a differentiable map representation built with 3D Gaussian Splatting (3DGS). 
Our key contribution is a geometry-aware training strategy that combines photometric and geometric losses, where the geometric supervision is provided for the first time by combining multi-view stereo (MVS) and LiDAR depths. We show that this joint optimization produces a 3DGS model that better fits the underlying scene geometry, leading to improved photometric and geometric consistency and more robust, accurate single-image 6-DoF pose estimation. Extensive experiments on data acquired in planetary-analog environments validate the effectiveness of our approach, showing substantial gains in relocalization accuracy under challenging conditions. 
Code is available at \url{https://github.com/DLR-RM/multimodal-gsplat-relocalization}. 



\end{abstract}

\section{INTRODUCTION}

\begin{figure}[t!]
    \centering
    \includegraphics[width=\linewidth]{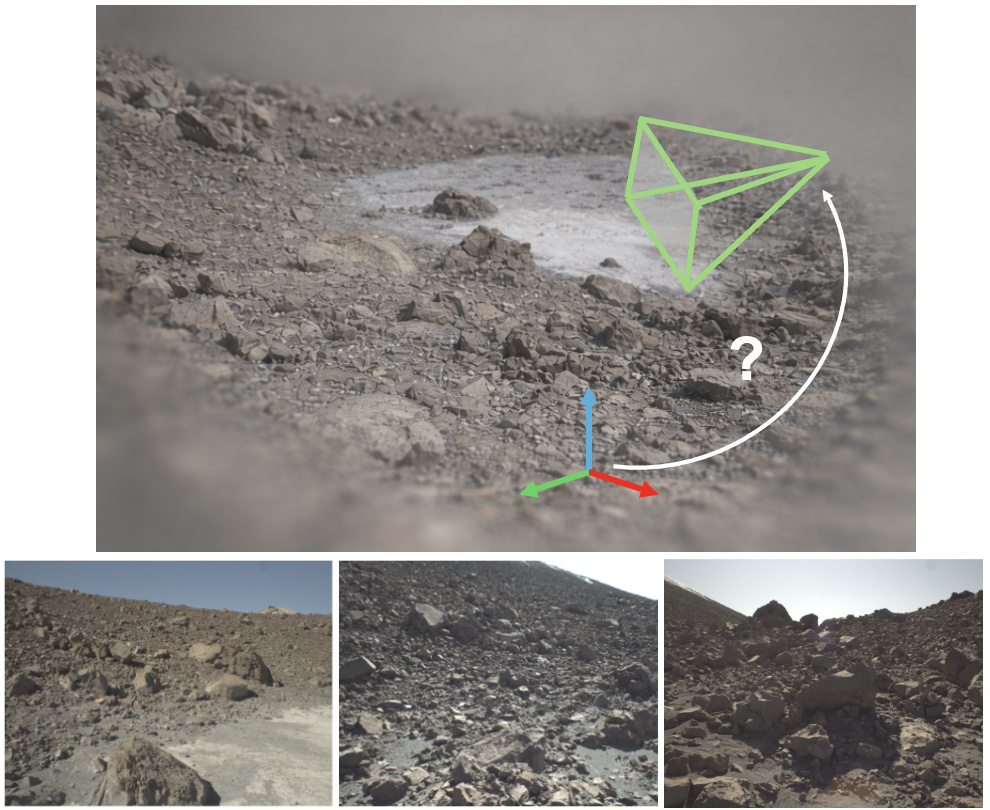}
    \caption{\textbf{Illustration of relocalization challenges in extreme, planetary-like environments.} Despite the three images at the bottom of the figure overlap, the large viewpoint change and weak, aliased textures pose substantial challenges for visual relocalization, i.e., estimating the six-degrees-of-freedom camera pose within the map shown at the top of the figure.}
    \label{fig:teaser}
\end{figure}

Reliable visual relocalization remains a major challenge for long-term robotic autonomy in GNSS-denied environments ~\cite{ebadi2023present}, such as planetary-like scenarios~\cite{geromichalos2020slam}, to enable loop closure and mitigate odometry drift. However, perceptual aliasing and harsh illumination significantly degrades the performance of traditional visual place recognition and pose estimation. Moreover, the predominantly forward motion of planetary rovers results in restricted viewpoint diversity, limited parallax between consecutive frames, and significant perspective-induced appearance changes over longer time intervals \cite{DLRdataset}. Fig.~\ref{fig:teaser} illustrates these aspects.

Advances in novel view synthesis (NVS) offer an alternative to feature-based localization~\cite{deng2025best,tosi2026nerfs}. Instead of matching salient image features, camera poses are estimated in an image-to-map manner against dense 3D scene representations with richer photometric modeling. In particular, 3D Gaussian Splatting (3DGS)~\cite{kerbl20233dgaussiansplattingrealtime} provides an explicit, differentiable representation that enables render-based alignment and improved robustness to viewpoint changes. However, standard photometric 3DGS training is ill-suited for rover-acquired outdoor sequences. First, predominantly forward motion yields weak multi-view constraints, leading to high reconstruction error. Second, the sparsity of viewpoints limits generalization. And third, wide baselines during revisits challenge naive photometric losses due to strong perspective distortions. Together, these factors degrade the quality of 3DGS representations and hinder reliable relocalization. 


In this work, we address these relocalization challenges arising in 3DGS-based scene representations at planetary environments with adverse perceptual conditions. Specifically, in addition to the original photometric loss, we propose explicit supervision of the geometry by 1) depth and normal alignment with multi-view stereo (MVS), and 2) a novel Chamfer-based alignment with LiDAR point clouds. To our knowledge, this is the first time that a combination of such losses has been proposed. We implement a full relocalization pipeline composed of a first coarse stage of visual place recognition, followed by a fine-grained six-degrees of freedom camera pose estimation within the 3DGS representation.




We evaluate the proposed pipeline on perceptually challenging data collected in a planetary-analogous environment~\cite{DLRdataset,giubilato2026s3li}. Our results show that enforcing geometric alignment significantly improves the fidelity of 3DGS representations and yields substantial gains in single-image relocalization accuracy. These findings highlight the critical role of high-quality geometric and photometric scene representations for robust robot localization in extreme environments.

\section{Related work}

\subsection{Visual relocalization} Visual SLAM pipelines typically address relocalization in two stages. First, retireving candidate images based on global visual appearance~\cite{schubert2023visual}, and then estimating fine-grained camera poses with geometric or learned methods. Classical feature-based approaches, such as ORB-SLAM~\cite{campos2021orb}, rely on bags of words of handcrafted features~\cite{galvez2012bags} followed by PnP~\cite{lepetit2009epnp} with RANSAC~\cite{fischler1981random} for robust outlier rejection. While effective in textured environments and under small viewpoint changes, these methods degrade quickly under wide baselines and in perceptually ambiguous scenes. Learned components significantly improve robustness in modern pipelines, e.g., DINOv2 SALAD~\cite{izquierdo2024optimaltransportaggregationvisual} and VGGT~\cite{wang2025vggt} in VGGT-SLAM~\cite{maggio2025vggt}. Nevertheless, accurate pose estimation still remains challenging in scenarios involving wide baselines and low-texture or weakly informative regions. LiDAR relocalization faces similar challenges in scenes with limited structure~\cite{cui2022bow3d,cattaneo2022lcdnet,shi2024fast}. Finally, while direct pose regression approaches such as~\cite{kendall2015posenet} are attractive due to their simplicity, they typically have lower accuracy than alternatives with explicit scene representations~\cite{sattler2019understanding}. 



\subsection{Novel View Synthesis} 
NVS techniques~\cite{tewari2022advances} estimate dense scene representations that can be directly aligned with new observations, offering a promising alternative for relocalization under significant viewpoint changes.
Early NVS methods relied on structure-from-motion inputs and view interpolation~\cite{chen1995view, seitz1996view}, which required densely sampled viewpoints and reliable correspondences. More recent approaches, such as Neural Radiance Fields (NeRF)~\cite{NeRF} model scenes as continuous functions optimized through differentiable volume rendering, achieving high visual fidelity at the cost of computational complexity and implicit geometry encoding. 
In constrast, 3D Gaussian Splatting (3DGS)~\cite{kerbl20233dgaussiansplattingrealtime} introduces an explicit, differentiable scene representation that enables real-time rendering and stable gradient-based optimization. However, 3DGS training remains largely driven by photometric supervision, which can lead to geometric instability in sparse-view outdoor scenarios. Only recently have several works incorporated either MVS~\cite{liu2024mvsgaussian,chen2024mvsplat,takama2025sparse2dgs} or LiDAR point clouds~\cite{zhou2024drivinggaussian,yan2024street, peng2025constrained} to improve geometric quality. To the best of our knowledge, ours is the first work to systematically demonstrate the complementarity of these supervision modalities and the performance gains obtained by combining both of them.

\subsection{Pose Estimation in Neural Representations} 
Neural scene representations have recently emerged as a compelling alternative to correspondence-based localization pipelines. Methods such as iNeRF~\cite{yen2020inerf} cast pose estimation as a render-and-compare optimization problem, directly aligning query images with a continuous scene model, but inheriting complexity and opaque representation from NERFs. 
Approaches such as 6DGS~\cite{bortolon20246dgs6dposeestimation} instead leverage Gaussian-based maps to estimate single-image 6-DoF poses without iterative photometric refinement, highlighting the promise of differentiable scene models for relocalization. Nevertheless, pose accuracy remains strongly dependent on the geometric and photometric fidelity of the underlying reconstruction, properties that often deteriorate in sparse-view, large-scale outdoor environments.

To address these limitations, recent works have incorporated LiDAR-based geometric priors into 3DGS to improve structural realism and geometric accuracy, including GS-LiDAR~\cite{jiang2025gslidargeneratingrealisticlidar} and SplatAD~\cite{hess2025splatadrealtimelidarcamera}. While these approaches enhance geometric alignment, their impact on downstream camera relocalization performance in challenging planetary-like scenarios remains largely unexplored. In contrast, our work further augments these geometric priors with MVS depths and normals, and explicitly investigates how geometry-aware Gaussian fitting affects single-image relocalization in sparse, forward-motion robotic settings.

\subsection{3D Gaussian Splatting for SLAM}
Recent works have begun integrating 3DGS into full SLAM pipelines, enabling real-time tracking and dense map construction using explicit splat-based scene representations. Gaussian-SLAM~\cite{matsuki2024gaussiansplattingslam} incorporates 3DGS within an online SLAM framework, jointly optimizing camera poses and Gaussian parameters for dense reconstruction and tracking. 

\begin{figure*}[ht!]
    \centering
    \includegraphics[width=\linewidth]{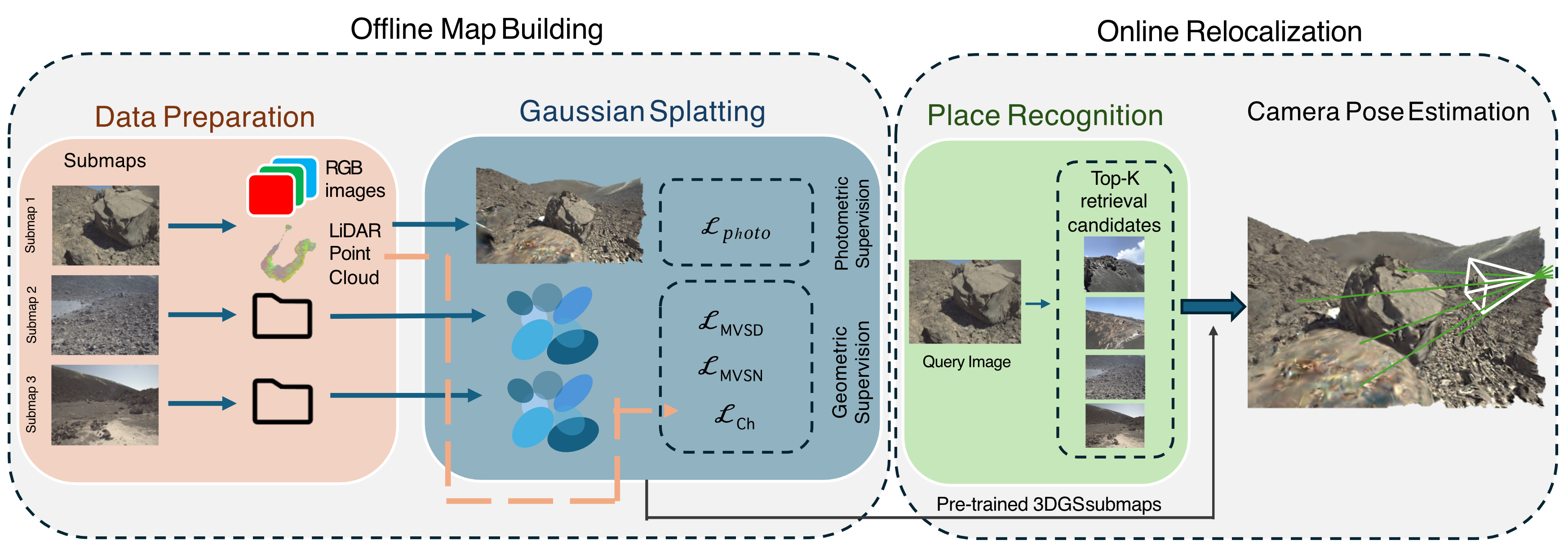}
    \caption{\textbf{Overview of our relocalization pipeline, based on structure-preserving 3DGS.} RGB images and LiDAR readings are grouped into submaps, and used to train a 3DGS with strong geometric supervision, in addition to the standard photometric supervision. At inference, the query image is used to retrieve top-k candidate submaps, and camera pose is estimated by minimizing the rendering error against the 3DGS representations.}
    \label{fig:pipeline}
\end{figure*}

Beyond tracking and mapping, LoopSplat~\cite{zhu2025_loopsplat} explores loop closure directly in the Gaussian domain by registering splat-based maps across revisited areas, enabling map-level alignment without relying on sparse feature correspondences. These approaches highlight the potential of explicit Gaussian representations for geometric alignment and global consistency.
Multi-modal extensions~\cite{sun2024mm3dgs} further integrate additional sensing modalities within Gaussian-based SLAM frameworks to enhance structural accuracy and robustness. However, existing GS-based SLAM systems primarily focus on tracking stability, reconstruction fidelity, or map registration strategies. The role of training-level geometric supervision and global consistency in improving downstream single-image relocalization and loop closure robustness, particularly under sparse forward motion and severe perceptual aliasing, remains largely unexplored.

In contrast, our work isolates and analyzes how geometry-aware training of Gaussian maps influences 6-DoF relocalization performance in challenging planetary-like environments, providing a controlled study of metric consistency for loop closure.

\section{Structure-Preserving Gaussian Splatting}\label{sec:map_representation}

\subsection{Gaussian Splatting Submaps}
\label{sec:gs_submaps}
Our overall scene representation $\mathcal{G}$ is composed of a set of $m$ submaps $\mathcal{G} = \{\mathcal{G}_1, \hdots, \mathcal{G}_s, \hdots, \mathcal{G}_m\}$, where each submap $\mathcal{G}_s = f(\mathcal{P}_s, \mathcal{I}_s, \mathcal{T}_s)$ models a part of the scene as a set of Gaussians encoding its geometry and visual appearance~\cite{kerbl20233dgaussiansplattingrealtime}. The Gaussian representation $\mathcal{G}_s$ will be estimated, as detailed in the next subsection, as a function $f$ of an aggregated LiDAR point cloud $\mathcal{P}_s = \{ \mathbf{P}_1^s, \hdots, \mathbf{P}_q^s, \hdots, \mathbf{P}_r^s \}$, $\mathbf{P}_j^s \in \mathbb{R}^3$, a set of $m$ RGB images $\mathcal{I}_s = \{\mathbf{I}_1^s, \hdots, \mathbf{I}_i^s, \hdots, \mathbf{I}_m^s\}$, $\mathbf{I}_i^s \in \mathbb{R}^{w \times h \times 3}$, and their respective camera poses $\mathcal{T}_s = \{\mathbf{T}_1^s, \hdots, \mathbf{T}_i^s, \hdots, \mathbf{T}_m^s\}$, $\mathbf{T}_i^s \in SE(3)$. 

Given this Gaussian Splatting-based scene representation $\mathcal{G}$ and a query image $\mathbf{I}_q$, visual relocalization consists in the estimation of the pose $\mathbf{T}_q^s$ of the query image in the reference frame of the corresponding submap. 
In a first stage, a set of candidate submaps is retrieved by visual place recognition among the submap images $\mathcal{I}_s$. After that, the 6-DoF pose $\mathbf{T}_q^s$ is estimated directly against the corresponding Gaussian representation. See Fig.~\ref{fig:pipeline} for an overview of the pipeline, and find additional details in the next subsections.


\subsection{Structure-Preserving 3DGS fitting}

3DGS fitting was originally formulated using only photometric supervision~\cite{kerbl20233dgaussiansplattingrealtime}, which is insufficient to ensure geometric fidelity, in particular in sparse-view outdoor scenarios. 
In this section, we first review the original photometric objective and then introduce our geometric supervision strategies: (i) depth- and normal-guided constraints derived from MVS to improve local surface coherence, and (ii) a LiDAR-guided Chamfer loss to enforce better global metric alignment.

\textbf{Photometric Supervision. }The photometric 3DGS loss typically combines a $\mathcal{L}_1$ loss and a structural similarity term
\begin{equation}
\mathcal{L}_{\mathrm{photo}} =
(1 - \lambda_{\mathrm{SSIM}})\mathcal{L}_1 + \lambda_{\mathrm{SSIM}} \mathcal{L}_{\mathrm{SSIM}}
\end{equation}

As in the original work~\cite{kerbl20233dgaussiansplattingrealtime}, we use $\lambda_{ssim} = 0.2$.

\textbf{MVSA Supervision. }
To mitigate the geometric ambiguities observed in the original 3DGS formulation, we introduce depth- and normal-guided supervision using MVSAnywhere (MVSA)~\cite{izquierdo2025mvsanywhere}, a multi-view stereo model with strong cross-domain generalization capabilities.
The predicted depth maps are incorporated as geometric supervision signals during training. Specifically, we introduce depth and surface normal consistency losses to further constrain the Gaussian parameters.
We enforce depth alignment through a $\mathcal{L}_1$ loss over valid pixels
\begin{equation}
\mathcal{L}_{\mathrm{MVSD}} =
\frac{1}{|\Omega|}
\sum_{i \in \Omega}
\left| d_i^{\mathrm{GS}} - d_i^{\mathrm{MVS}} \right|~,
\end{equation}
where \(d^{\mathrm{GS}}\) denotes the depth rendered from the 3DGS model, \(d^{\mathrm{MVS}}\) the depth predicted by MVSA, and \(\Omega\) the set of valid pixels.

Additionally, surface normal consistency is enforced through a cosine similarity loss
\begin{equation}
\mathcal{L}_{\mathrm{MSVN}} =
1 -
\frac{1}{|\Omega|}
\sum_{i \in \Omega}
{\mathbf{n}_i^{\mathrm{GS}}}^\top \cdot \mathbf{n}_i^{\mathrm{MVS}}~,
\end{equation}

\noindent where surface normals are obtained by differentiating the depth maps predicted by MVSAnywhere.

Together, these geometric constraints enhance local surface coherence and alleviate depth ambiguities in weakly textured regions, resulting in a more stable and structurally consistent reconstruction. Nevertheless, since multi-view depth estimation fundamentally depends on geometric parallax, its reliability decreases under sparse viewpoints and near-linear rover trajectories. In such scenarios, the limited baseline leads to noisy or biased depth predictions, particularly in distant or poorly observed areas.

\textbf{LiDAR Supervision. }
To further enforce global geometric consistency, we introduce a LiDAR-guided Chamfer-based loss that aligns the reconstructed Gaussian scene with metrically accurate LiDAR measurements.
Given the predicted point cloud $\mathcal{P}_{\mathrm{GS}}$ obtained from rendered depth maps and the LiDAR point cloud $\mathcal{P}_s$, we define a symmetric Chamfer loss
\begin{equation}
\mathcal{L}_{\mathrm{Ch}} =
\mathcal{L}_{\mathrm{acc}} +
\lambda_{\mathrm{comp}} \mathcal{L}_{\mathrm{comp}}~,
\end{equation}
\noindent \noindent where the weighting factor $\lambda_{\mathrm{comp}}$ balances both objectives. $\mathcal{L}_{\mathrm{acc}}$ measures reconstruction accuracy and $\mathcal{L}_{\mathrm{comp}}$ enforces geometric completeness and are formulated as follows
\begin{equation}
\mathcal{L}_{\mathrm{acc}} =
\frac{1}{|\mathcal{P}_{\mathrm{GS}}|}
\sum_{\mathbf{p} \in \mathcal{P}_{\mathrm{GS}}}
\min_{\mathbf{q} \in \mathcal{P}_s}
\|\mathbf{p} - \mathbf{q}\|_2 
\end{equation}
\begin{equation}
\mathcal{L}_{\mathrm{comp}} =
\frac{1}{|\mathcal{P}_s|}
\sum_{\mathbf{q} \in \mathcal{P}_s}
\min_{\mathbf{p} \in \mathcal{P}_{\mathrm{GS}}}
\|\mathbf{q} - \mathbf{p}\|_2~.
\end{equation}

This formulation is particularly suited to our setting, as it tolerates large density differences between sparse LiDAR measurements and dense Gaussian reconstructions, requires no explicit correspondences, and remains fully differentiable. By anchoring the Gaussian scene to metrically accurate LiDAR observations, the proposed loss resolves depth at low parallax and directly improves the stability and accuracy of downstream 6-DoF pose estimation.

The overall loss function $\mathcal{L}$ is defined as the weighted sum of the losses defined before
\begin{equation}
\mathcal{L} =
\ \mathcal{L}_{\mathrm{photo}}
+ \lambda_{\mathrm{MVSD}}\mathcal{L}_{\mathrm{MVSD}} 
+ \lambda_{\mathrm{MVSN}}\mathcal{L}_{\mathrm{MVSN}}
+ \lambda_{\mathrm{Ch}}\mathcal{L}_{\mathrm{Ch}}
\end{equation}

\section{relocalization from Geometry-Aware 3DGS}
\label{sec:vpr_pose_est}

Given the query image $\mathbf{I}_q$ and our 3DGS representation $\mathcal{G}$ described above, we aim to estimate the camera pose $\mathbf{T}_q^s$ in the reference frame of the corresponding submap $\mathcal{G}_s$.
To reduce the search space, we first perform visual place recognition to retrieve a set of candidate submaps. We adopt an existing pipeline~\cite{alejandra2025multi} built upon SALAD \cite{izquierdo2024optimaltransportaggregationvisual}, and use FAISS indexing \cite{douze2025faiss} to produce a coarse top-$k$ retrieval of similarly-looking images.

With the image candidates obtained from visual place recognition, we employ the 6DGS framework~\cite{bortolon20246dgs6dposeestimation}, which estimates the 6-DoF camera pose directly from a fixed 3DGS representation using feature-based ray–pixel correspondences and a weighted least-squares formulation.
The 6DGS pipeline is used without modification. Across all experiments, the pose estimation algorithm remains fixed, and only the training strategy of the underlying 3DGS map is varied. 
This controlled setup enables us to isolate the effect of the map's geometric and photometric fidelity on relocalization accuracy.


\section{Experiments}

\subsection{Submap Data Generation and Test Dataset}
\begin{figure}[!t]
    \centering
    \includegraphics[width=\linewidth]{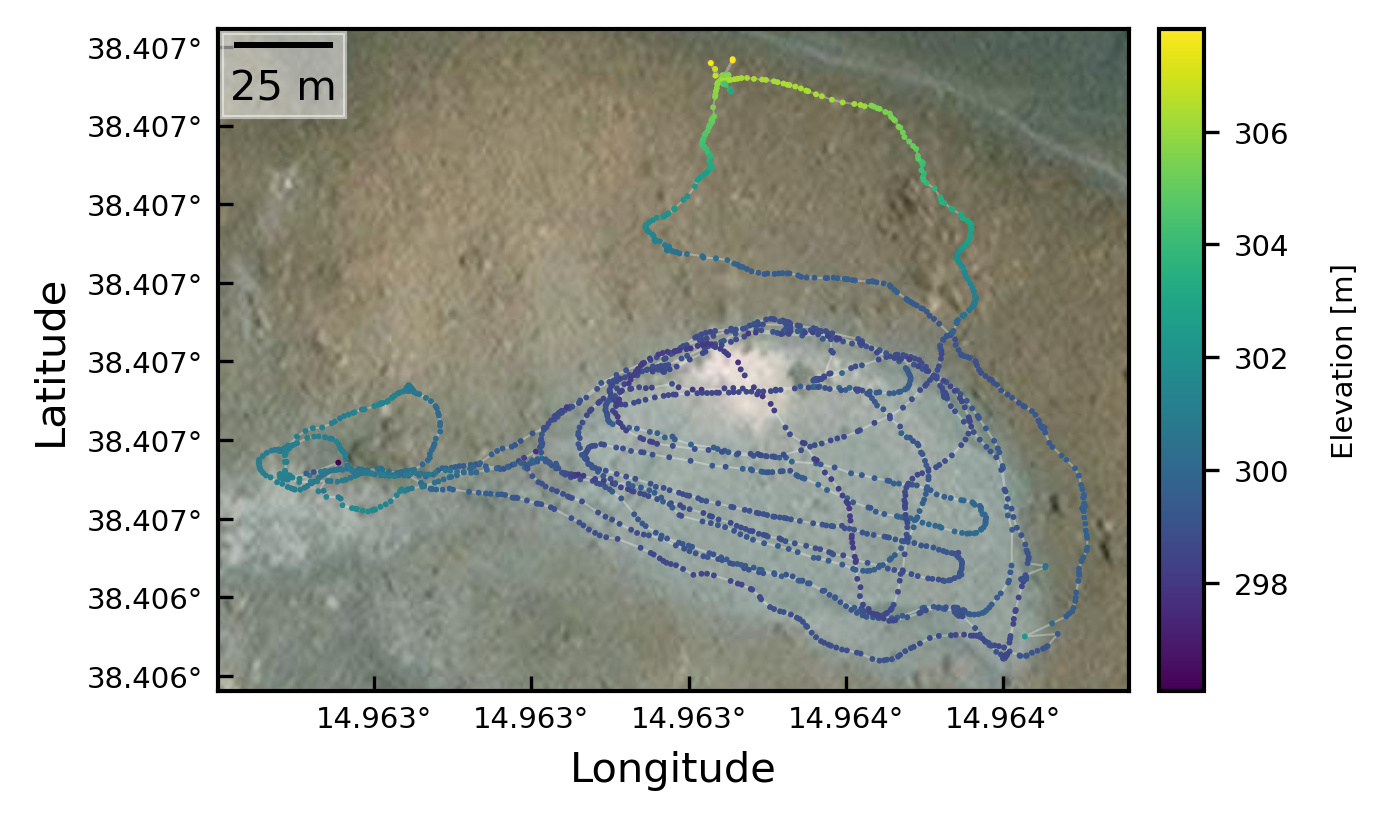}
    \caption{\textbf{Top-down view of the \textit{moon\_lake} sequence used in this work}, from the DLR S3LI Vulcano dataset. Impressions from the sequence are presented in Fig.~\ref{fig:teaser}}.
    \label{fig:moon_lake}
\end{figure}
We rely on a multi-modal submap-based SLAM approach \cite{giubilato2021multi} to generate the submaps containing LiDAR point clouds $\mathcal{P}_s$, RGB images $\mathcal{I}_s$ and respective poses $\mathbf{T}_i^s$, as defined in Sec.~\ref{sec:gs_submaps}. A visual-inertial front-end provides local pose estimates with respect to switching reference frames defining the origin of each submap. RGB images refer to visual keyframes extracted using a traditional co-visibility policy, with their pose in the parent submap reference frame. LiDAR point clouds are aggregated based on local VIO estimates.
Intrinsic camera parameters and keyframe poses are fixed during training, and geometry-aware variants of the 3DGS training pipelines operate directly in metric scale.

Focusing on the task of relocalization on challenging unstructured terrains, for which available data is extremely scarce, we test our proposed approach on the DLR S3LI Vulcano Dataset~\cite{giubilato2026s3li}. The dataset targets multi-modal SLAM and place recognition recorded on planetary analogous terrains from the perspective of a mobile rover. The sequences we focus on belong to an extension of the original dataset~\cite{DLRdataset} and take place on the island of Vulcano, Sicily, where an old inactive crater creates a valuable playground to test the limits of place recognition. The data consists of synchronized RGB images and LiDAR point clouds, and differential GNSS ground truth. 
In this work we focus on the \textit{moon\_lake} sequence from the dataset, offering a challenging benchmark for place recognition tasks. The sequence spans about 25 minutes of recordings of multi-modal data along a trajectory of 1.5 km length. A top-down view of d-GNSS data points and elevation is shown in Fig.~\ref{fig:moon_lake}. The trajectory is characterized by predominantly forward-facing motion with limited viewpoint diversity, resulting in limited parallax among observations of significant points of interest. 
Repeated observations of previously visited places happen without specific patterns of viewpoint intersection, as the observer is unconstrained in traversing the environment. 
Furthermore, the absence of man-made features, such as hiking paths or visual landmarks, cause significant perceptual aliasing. 

\subsection{Baselines}

We compare the proposed geometry-aware 3DGS representation against two baselines:

\noindent\textbf{3DGS.}
As a direct baseline, we train standard 3D Gaussian Splatting models using only photometric supervision, without any geometric constraints. The retrieval stage and pose estimation pipeline (6DGS~\cite{bortolon20246dgs6dposeestimation}), introduced in Sec.~\ref{sec:vpr_pose_est}, remain identical to our method. 

\noindent\textbf{PnP.}
For comparison with classical relocalization pipelines, we implement a feature-based PnP baseline. For each query image, the top-$K$ candidate submaps are retrieved using the same place recognition pipeline. SuperPoint features~\cite{detone2018superpoint} are extracted from the query and retrieved images, and are matched with SuperGlue~\cite{sarlin2020supergluelearningfeaturematching}. Using the depth map associated with each retrieved image, matched keypoints are back-projected to obtain 3D points. The camera pose of the query image is then estimated from the resulting 2D--3D correspondences using a RANSAC-based PnP solver.

\subsection{Training Details}

All experiments are conducted using the Nerfstudio framework. 
Photo-3DGS models are trained independently for each submap using the standard \textit{splatfacto} pipeline for $30{,}000$ iterations with the Adam optimizer. 
Camera intrinsics and poses are kept fixed, and Nerfstudio’s built-in pose normalization and scene centering mechanisms are enabled. 
No explicit geometric supervision is applied beyond photometric consistency for Photo-3DGS. Geometry-aware variants are trained using the \textit{regsplatfacto} pipeline, which incorporates additional geometric regularization terms. In contrast to the baseline, automatic pose normalization and scene centering are disabled, and optimization is performed directly in the metric reference frame. 
Each submap is trained for $20{,}000$ iterations with identical optimizer settings.
Depth and surface normal supervision are provided by MVSAnywhere, initialized from the pretrained checkpoint \texttt{mvsanywhere\_hero.ckpt}.
The optimization objective combines photometric loss with additional depth and normal regularization terms weighted by $\lambda_{\mathrm{MVSD}} = 0.05$ and $\lambda_{\mathrm{MVSN}} = 0.1$.
For the final variant, a LiDAR-guided Chamfer loss is introduced to enforce global metric scale consistency.
The Chamfer term is activated after $2{,}000$ iterations and linearly increased until iteration $8{,}000$, reaching a final weight of $\lambda_{\mathrm{Ch}} = 5 \times 10^{-5}$. All remaining hyperparameters are kept fixed to isolate the impact of geometric supervision.

\subsection{Reconstruction Results}

We begin by evaluating the geometric fidelity of the reconstructed 3DGS models, as reconstruction quality has a direct impact on downstream relocalization accuracy.

\begin{figure*}[t]
\centering
\setlength{\tabcolsep}{1pt}
\renewcommand{\arraystretch}{1.0}

\newcommand{\rowlabel}[1]{%
    \rotatebox{90}{\parbox{1cm}{\centering\textbf{#1}}}
}
\newcommand{\rowstrut}{\rule{0pt}{1.3cm}}

\newcommand{\img}[1]{%
    \includegraphics[width=0.13\linewidth]{#1}
}

\newcommand{\imgGT}[1]{%
    \includegraphics[height=1.33cm,keepaspectratio]{#1}
}

\newcommand{\imgGTtwo}[1]{%
    \includegraphics[height=1.2cm,keepaspectratio]{#1}
}

\begin{tabular}{c ccccccc}
&
{\small\textbf{GT}} &
{\small\textbf{3DGS}} &
{\small\textbf{\shortstack{3DGS\\+ LiDAR}}} &
{\small\textbf{\shortstack{3DGS\\+ MVSA}}} &
{\small\textbf{\shortstack{3DGS\\+ MVSA\\+ LiDAR}}} &
{\small\textbf{\shortstack{3DGS\\+ Chamfer\\+ LiDAR}}} &
{\small\textbf{Ours}} \\
\addlinespace[0.2mm]


\rowlabel{RGB}\rowstrut &
\imgGT{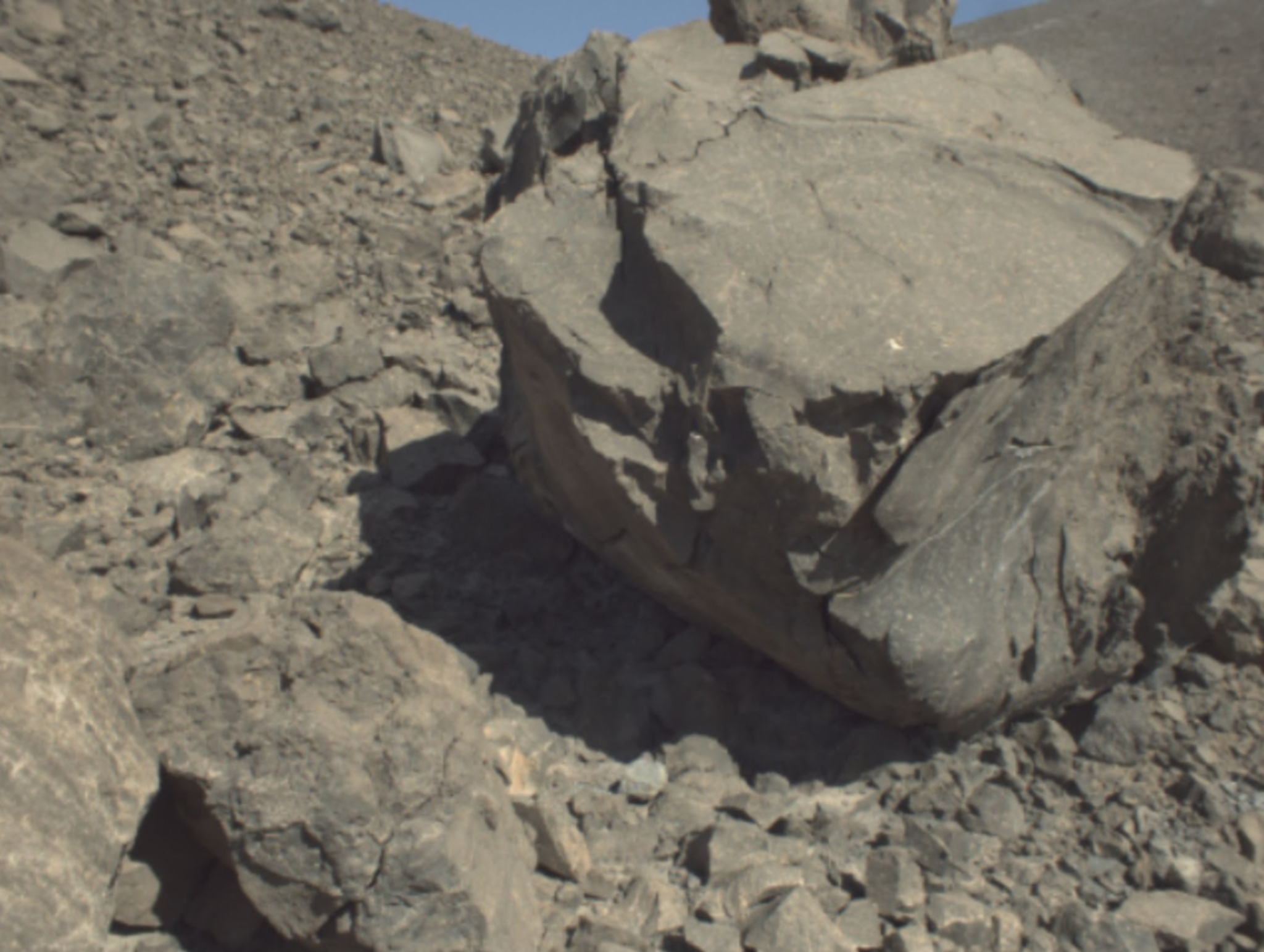} &
\img{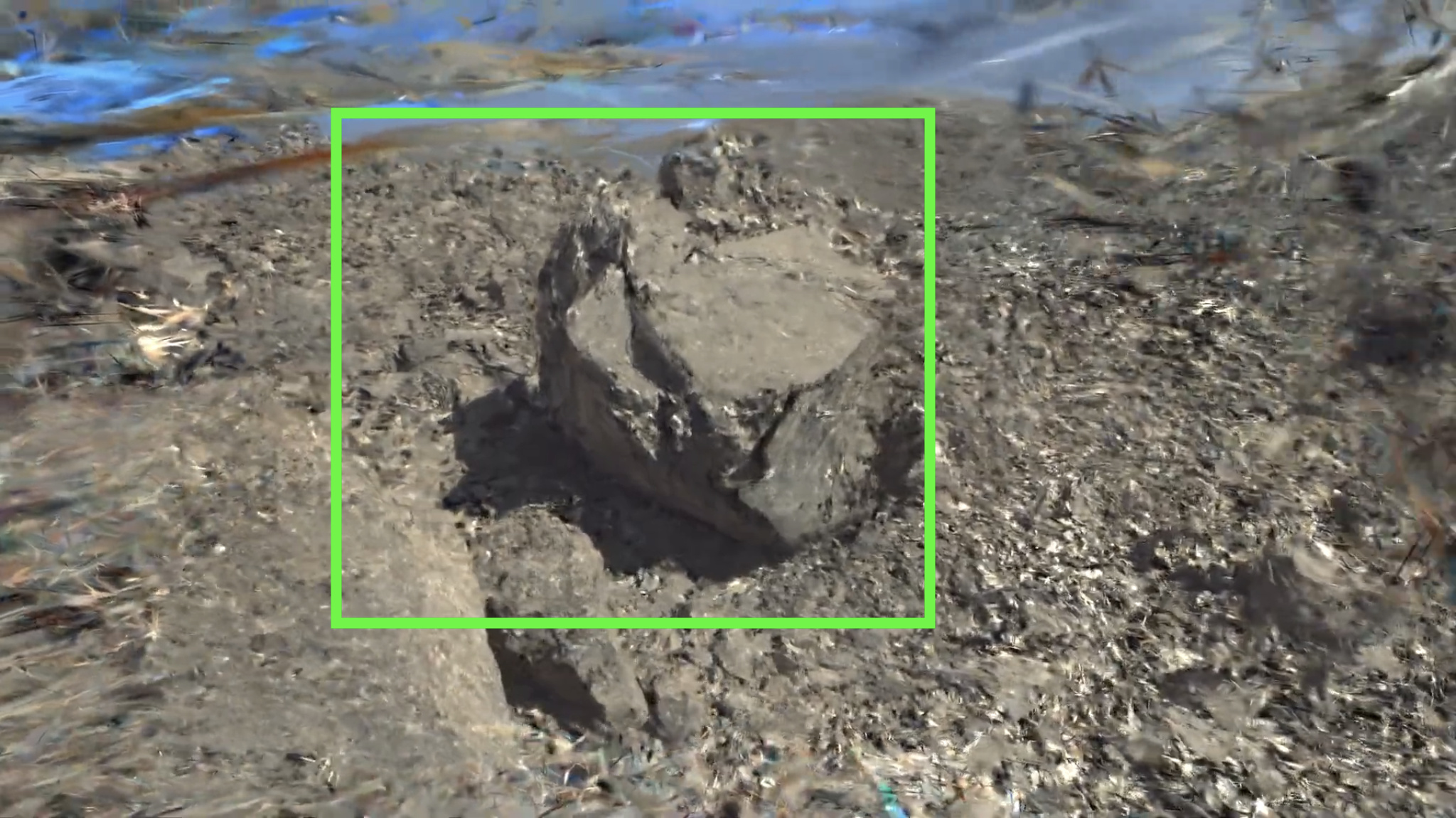} &
\img{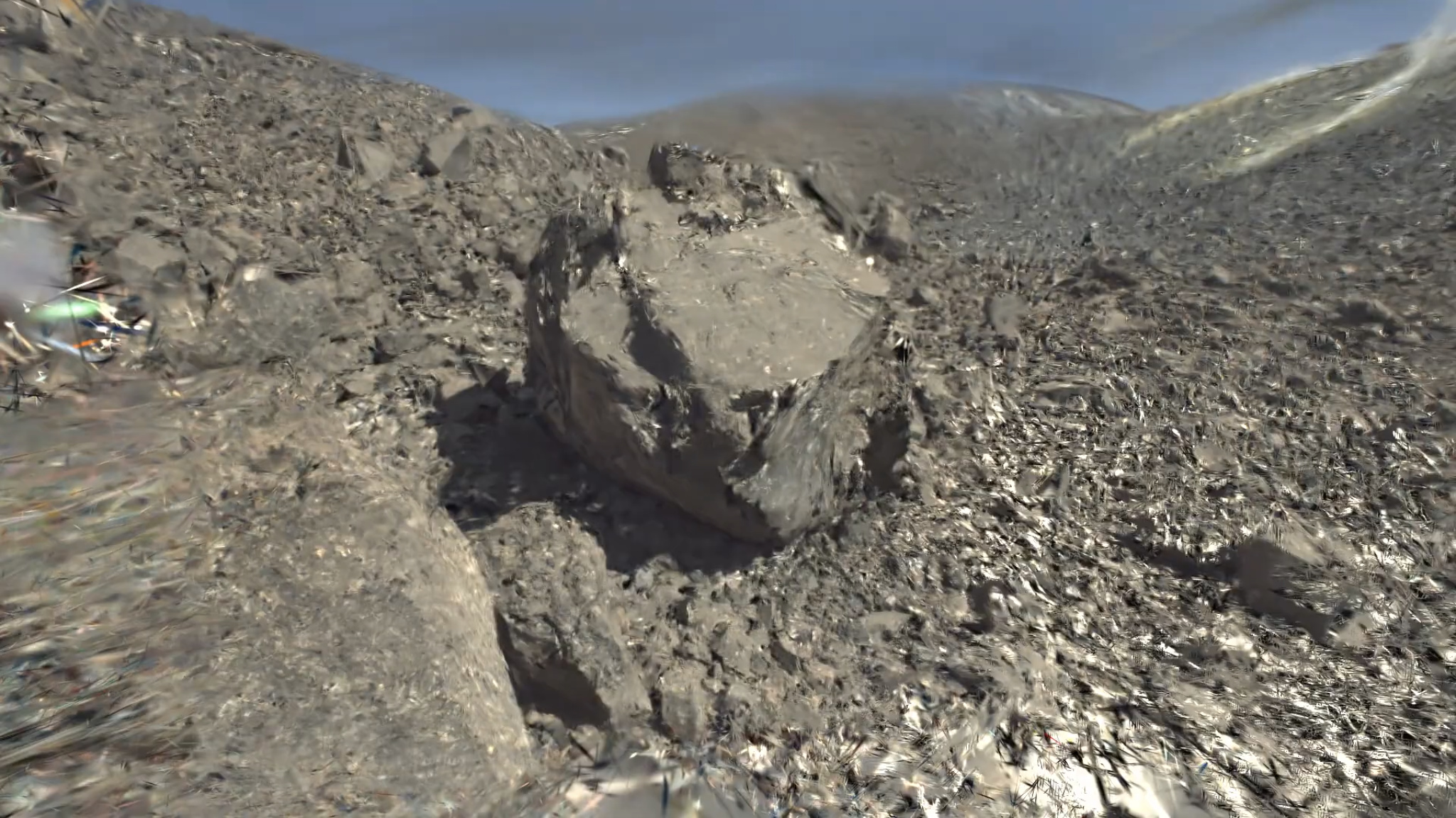} &
\img{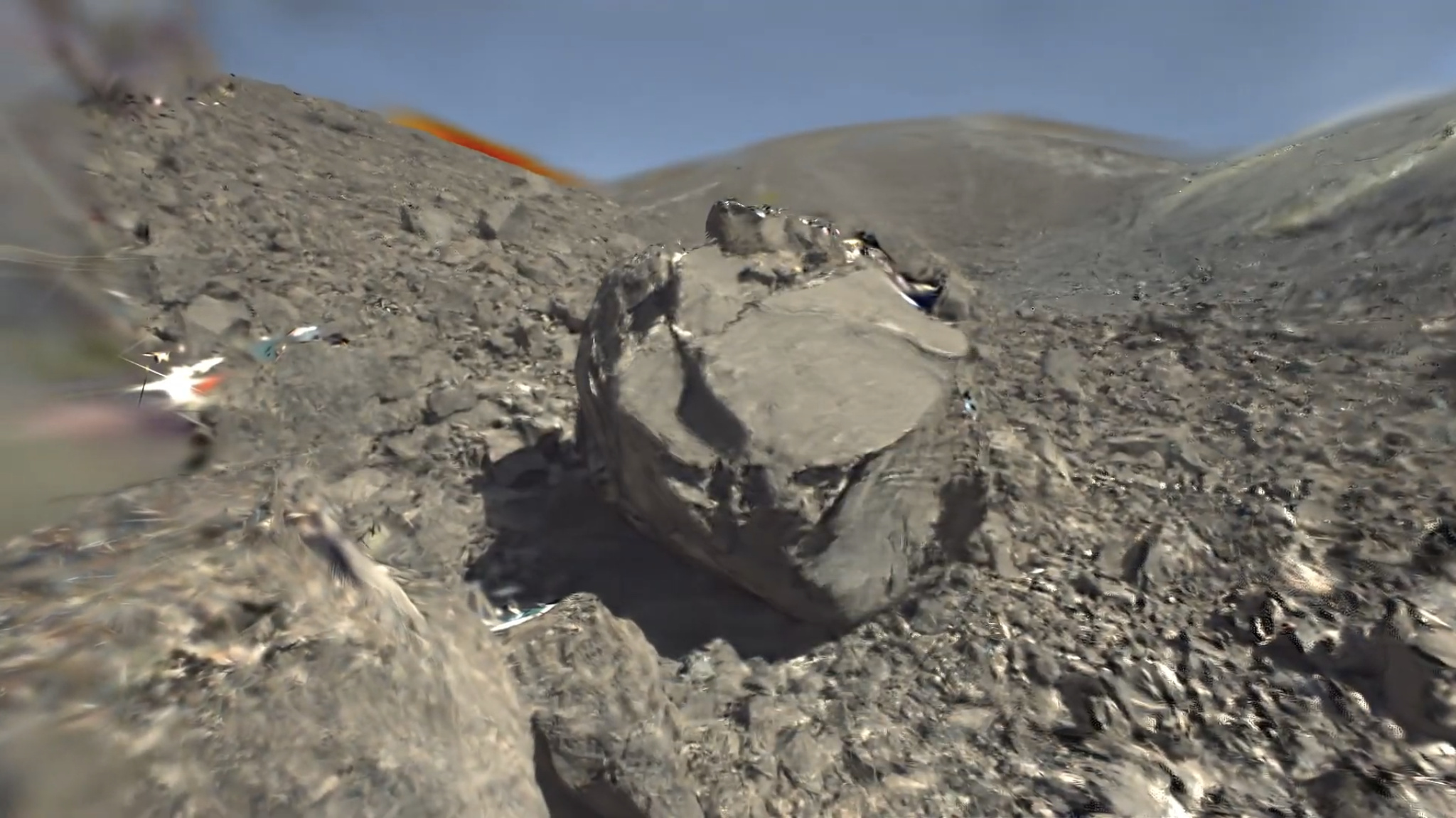} &
\img{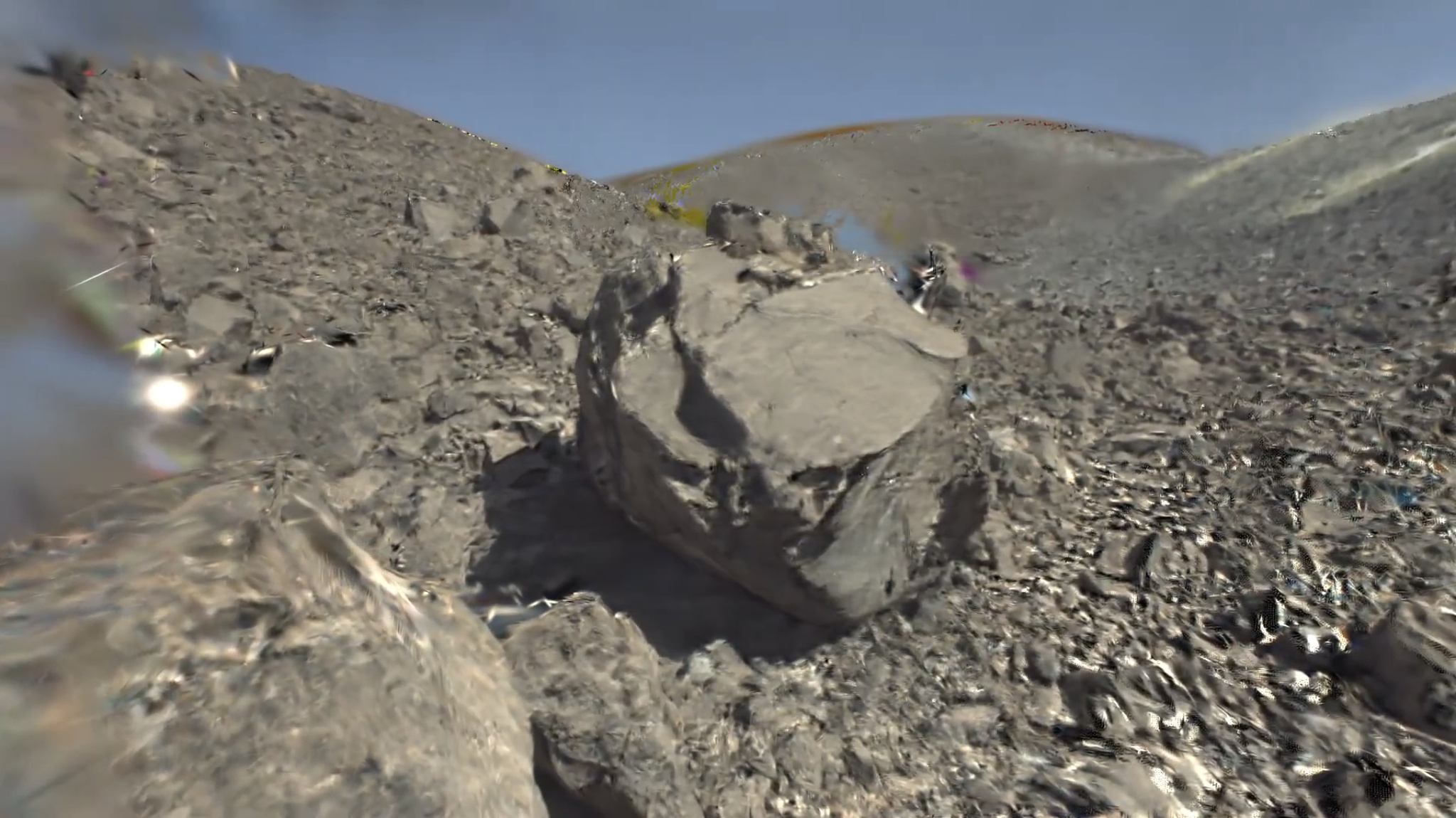} &
\includegraphics[width=0.13\linewidth,trim=150mm 50mm 50mm 50mm,clip]{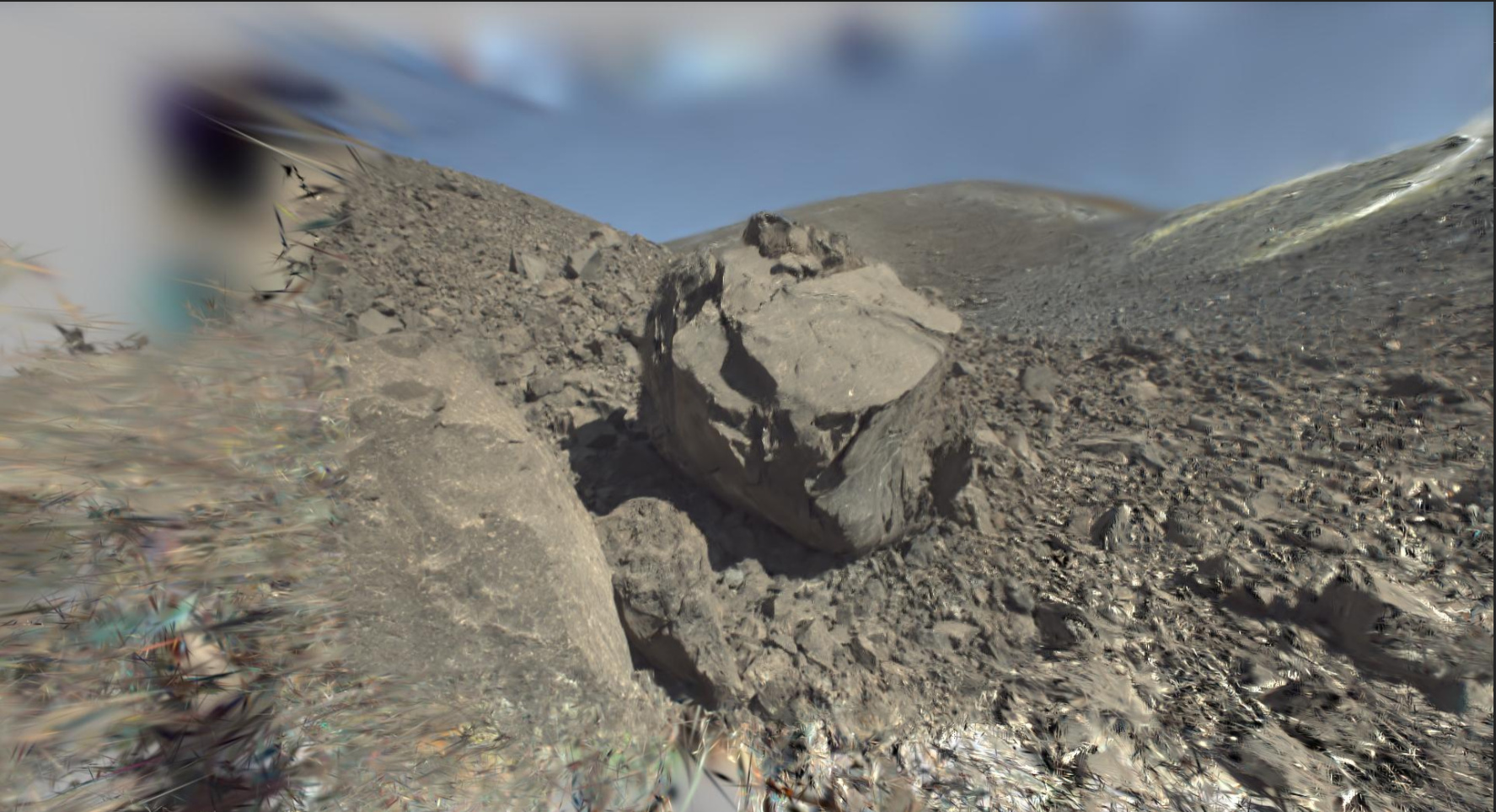} &
\img{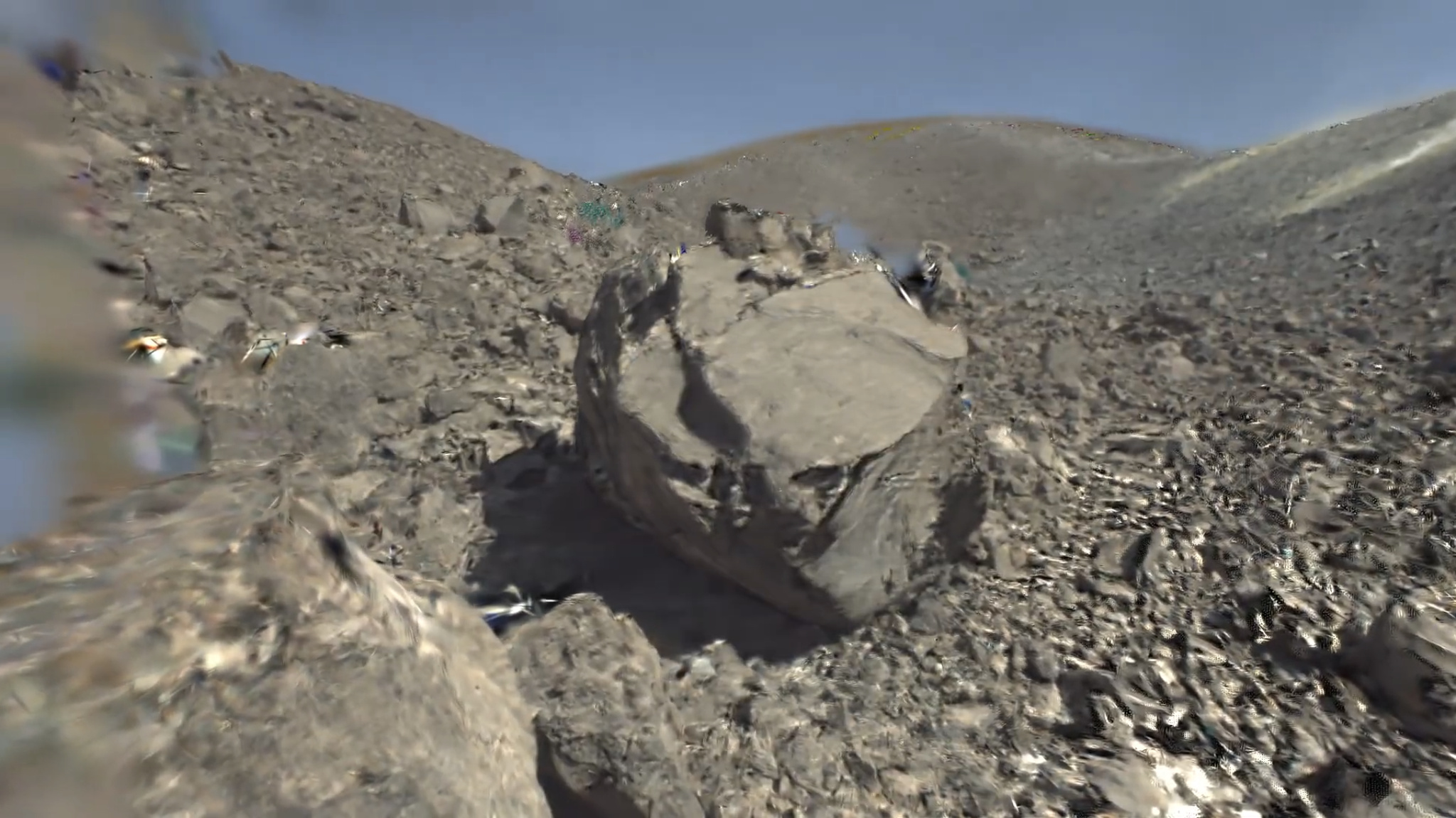} \\[-0.05mm]

\rowlabel{Depth}\rowstrut &
&
\img{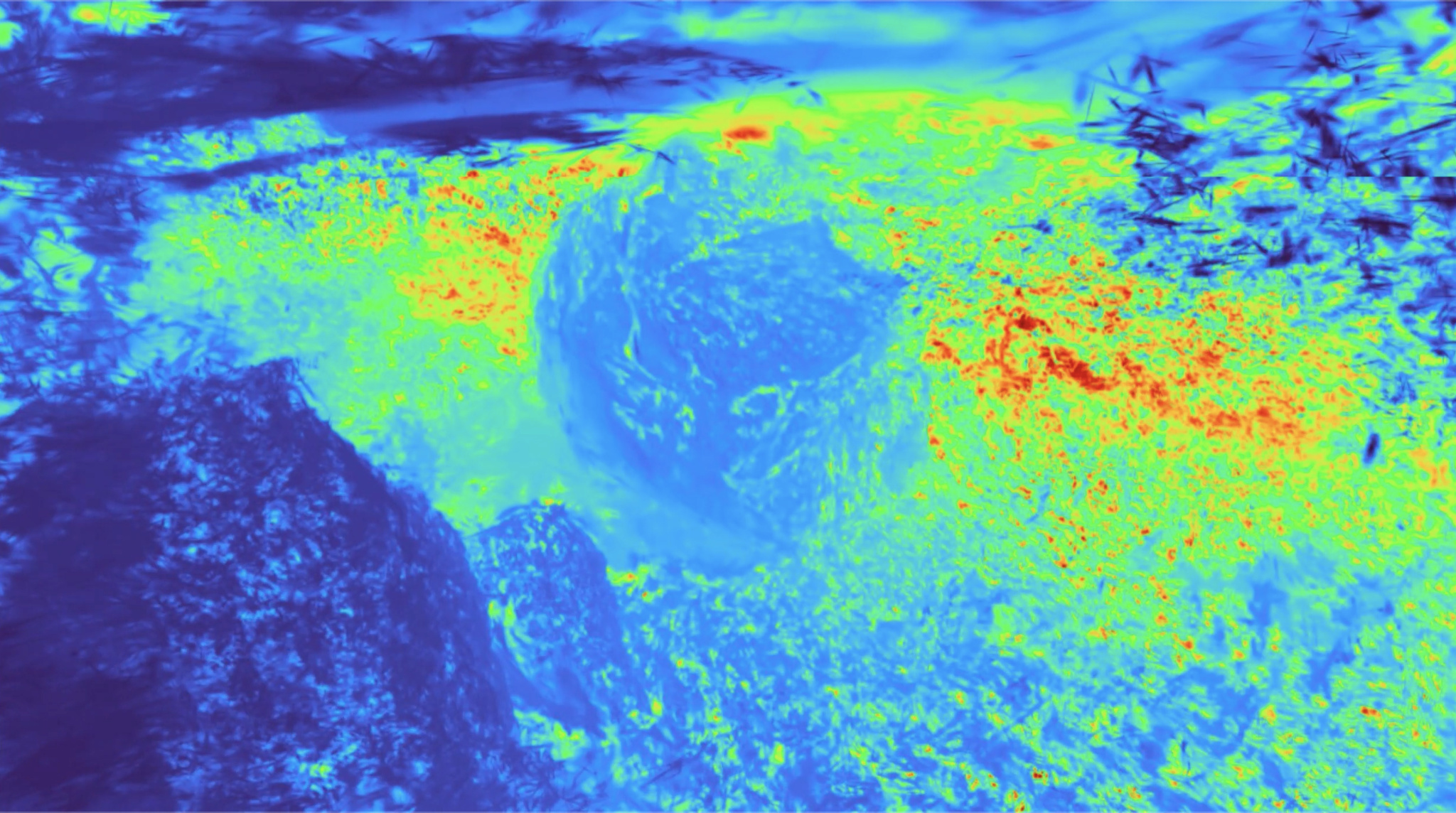} &
\img{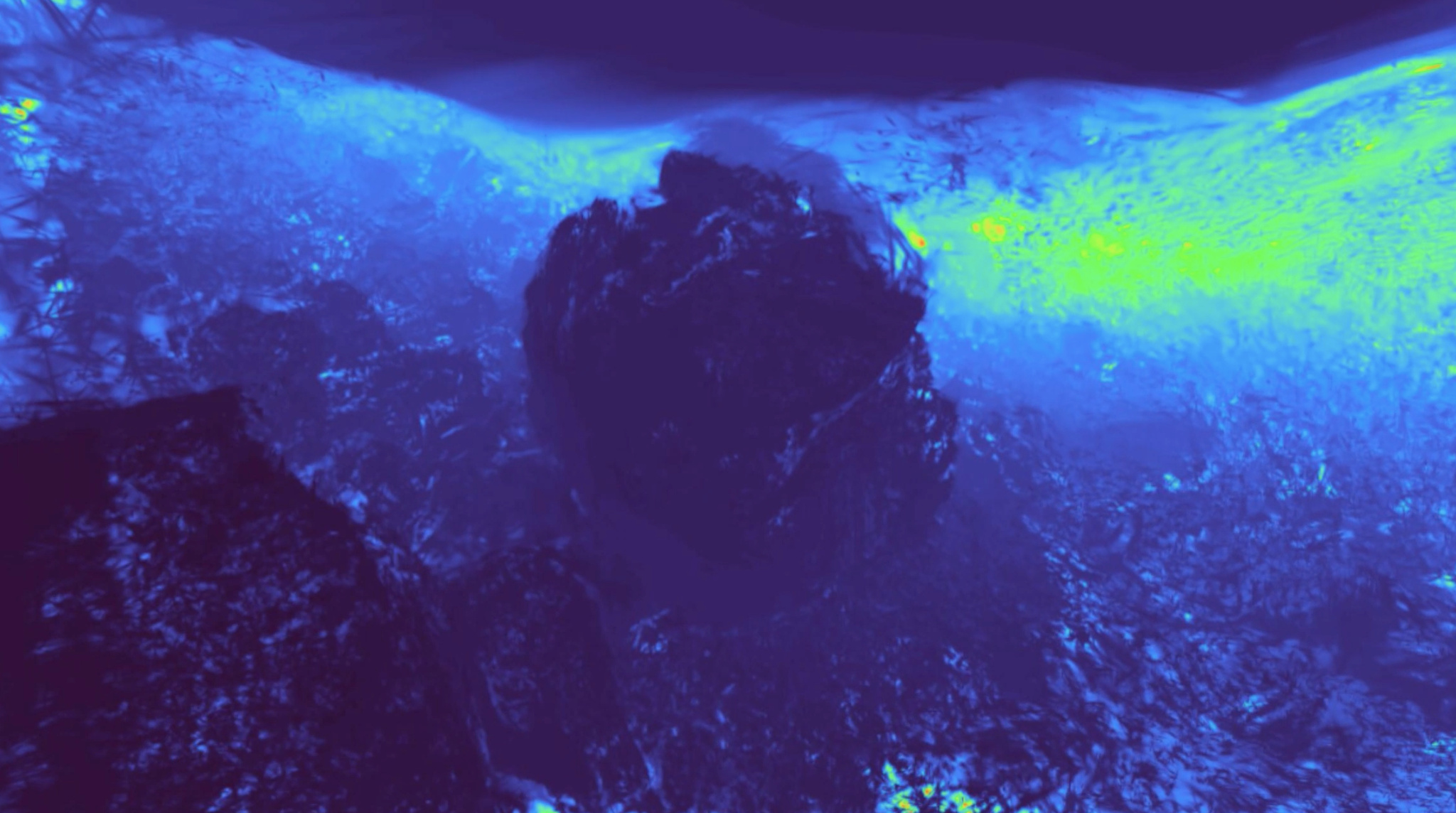} &
\img{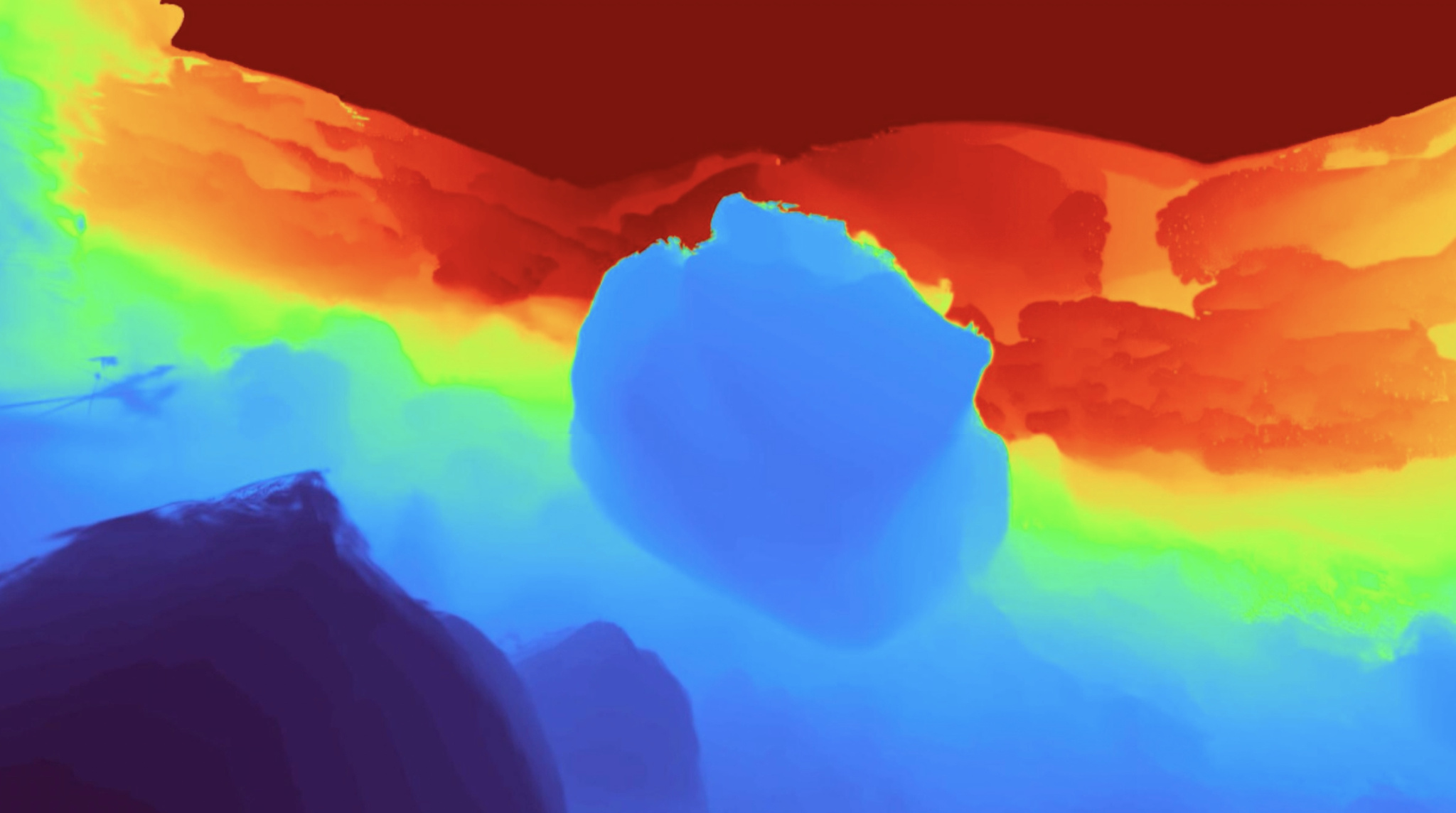} &
\img{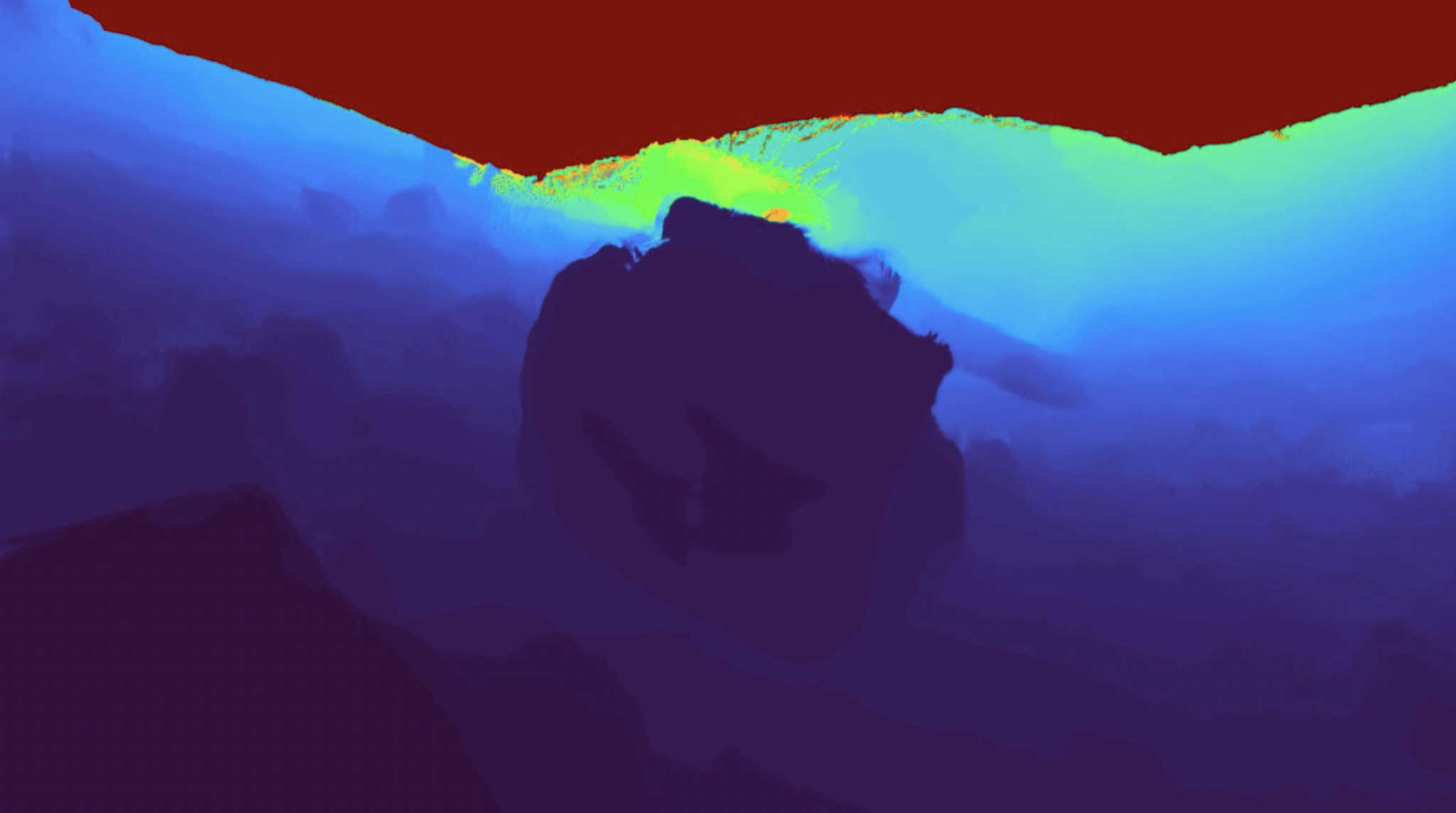} &
\includegraphics[width=0.13\linewidth,trim=150mm 50mm 50mm 50mm,clip]{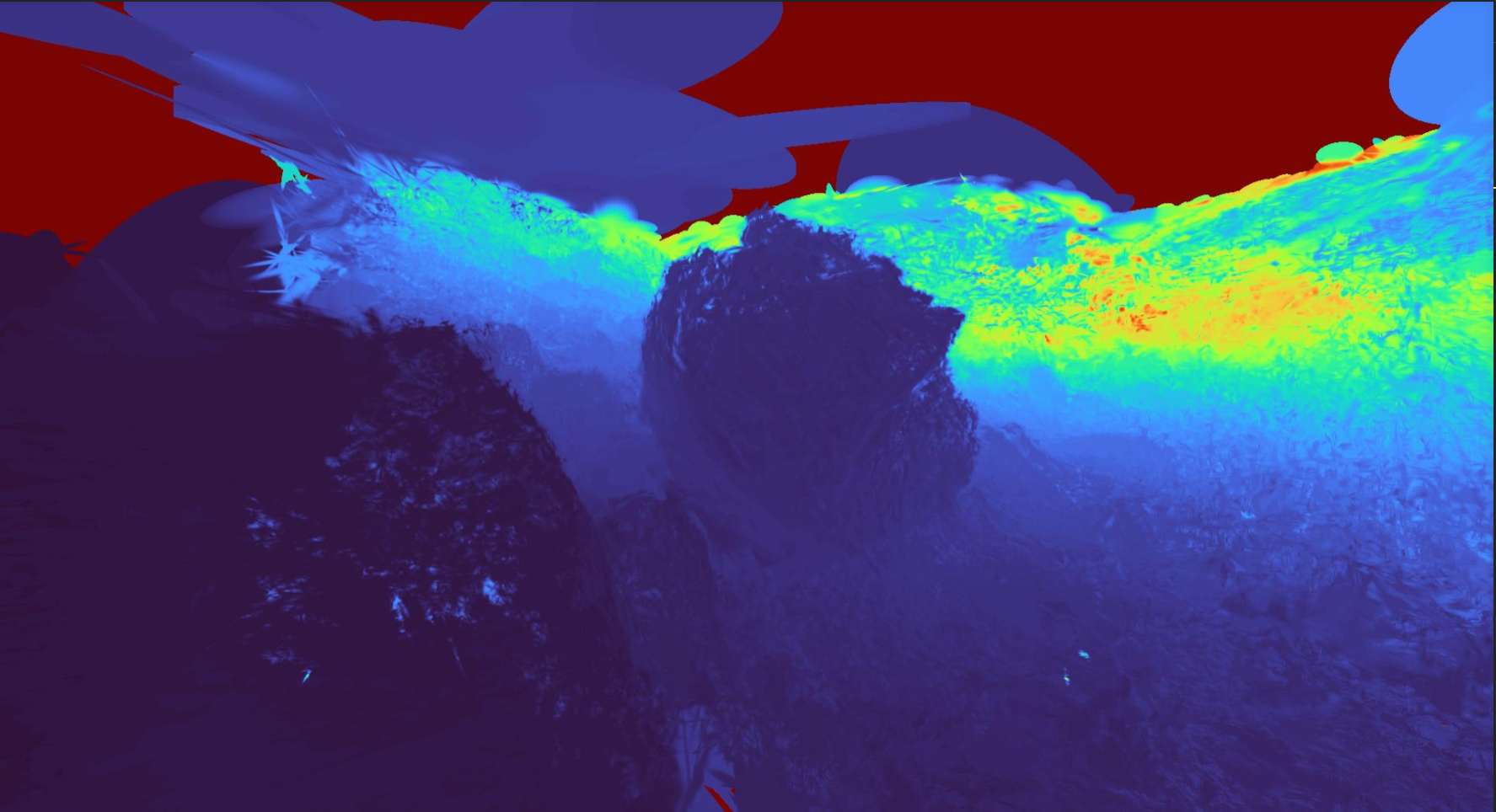} &
\img{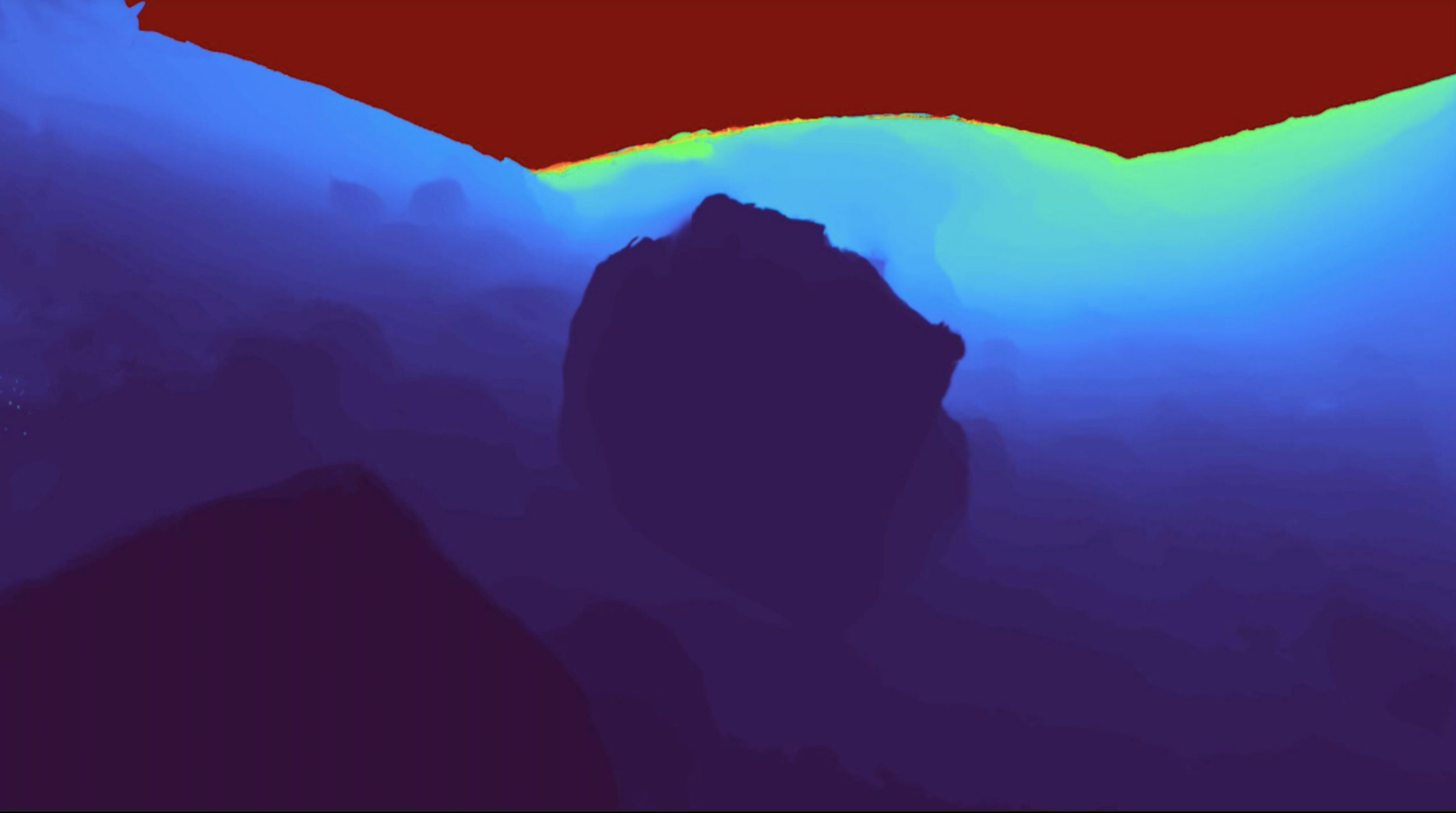} \\


\rowlabel{RGB}\rowstrut &
\imgGTtwo{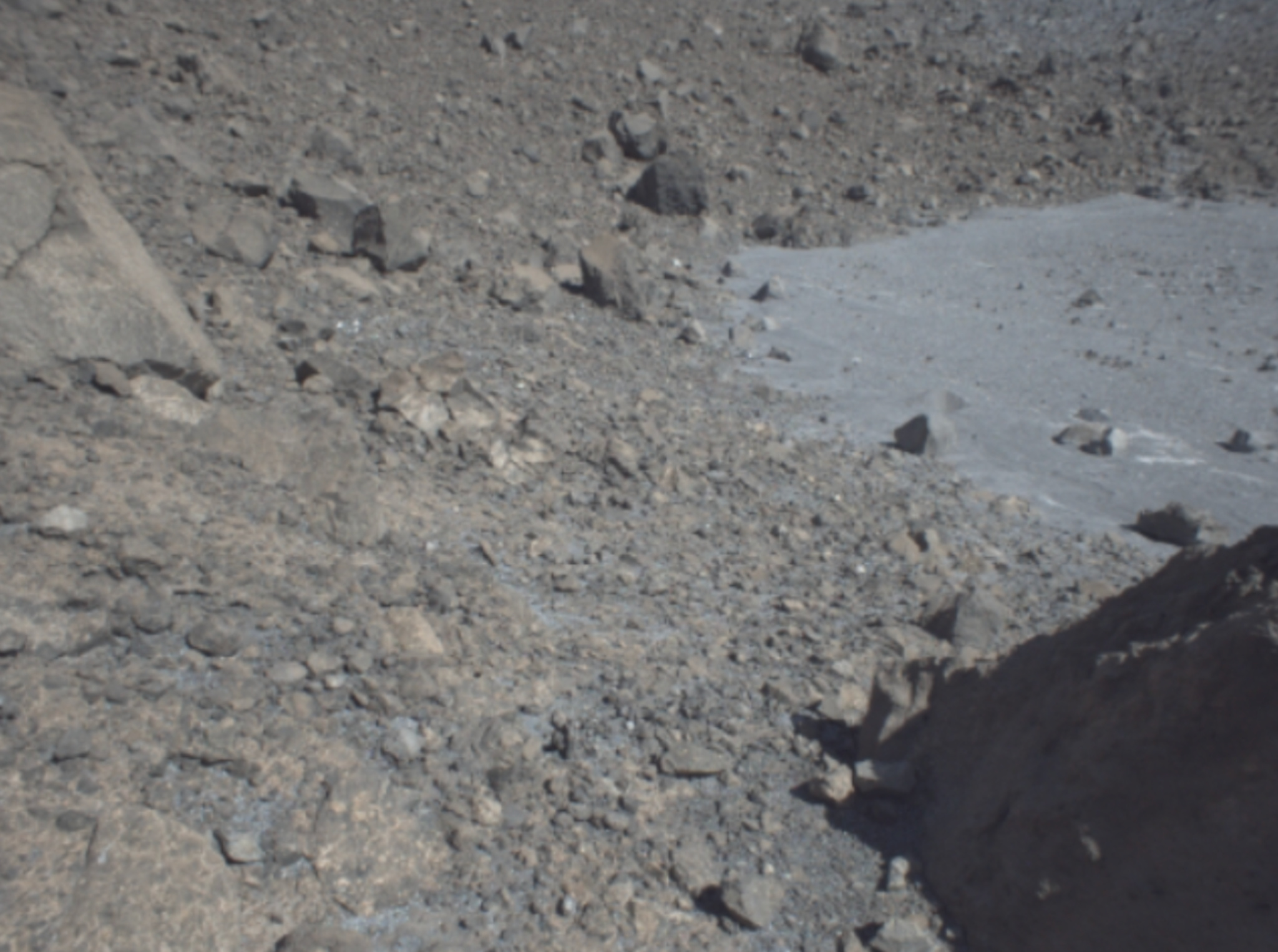} &
\img{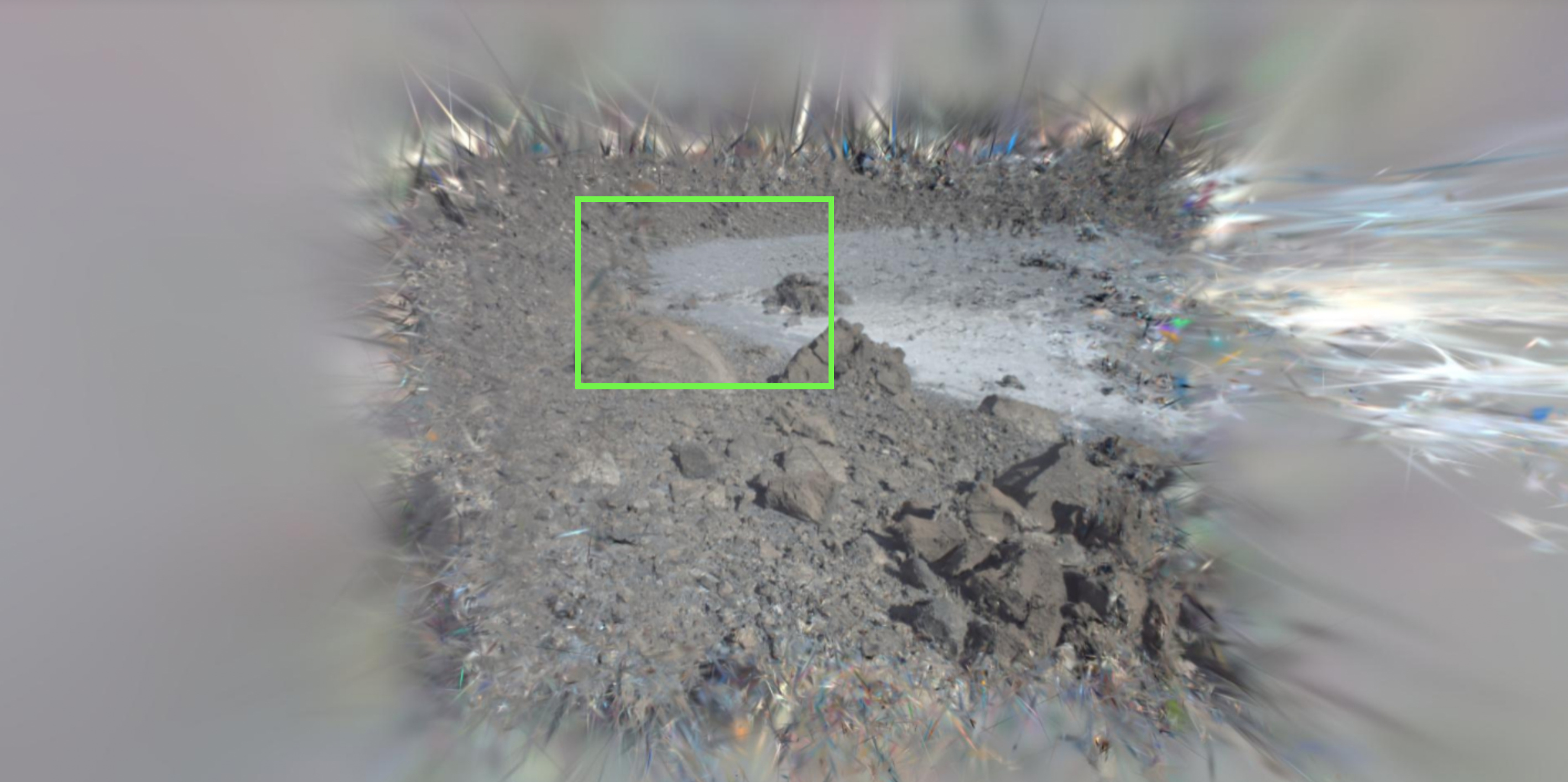} &
\img{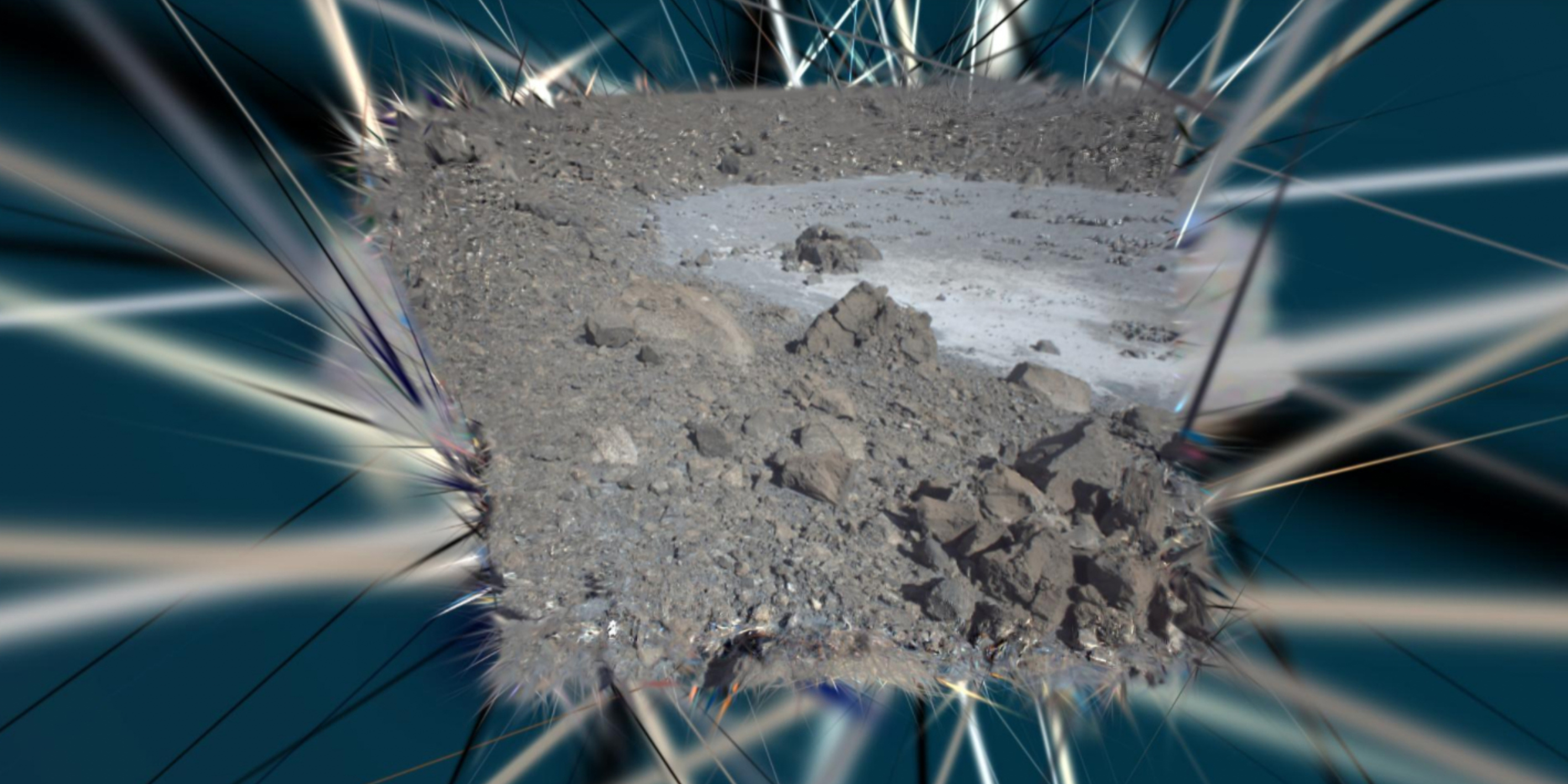} &
\img{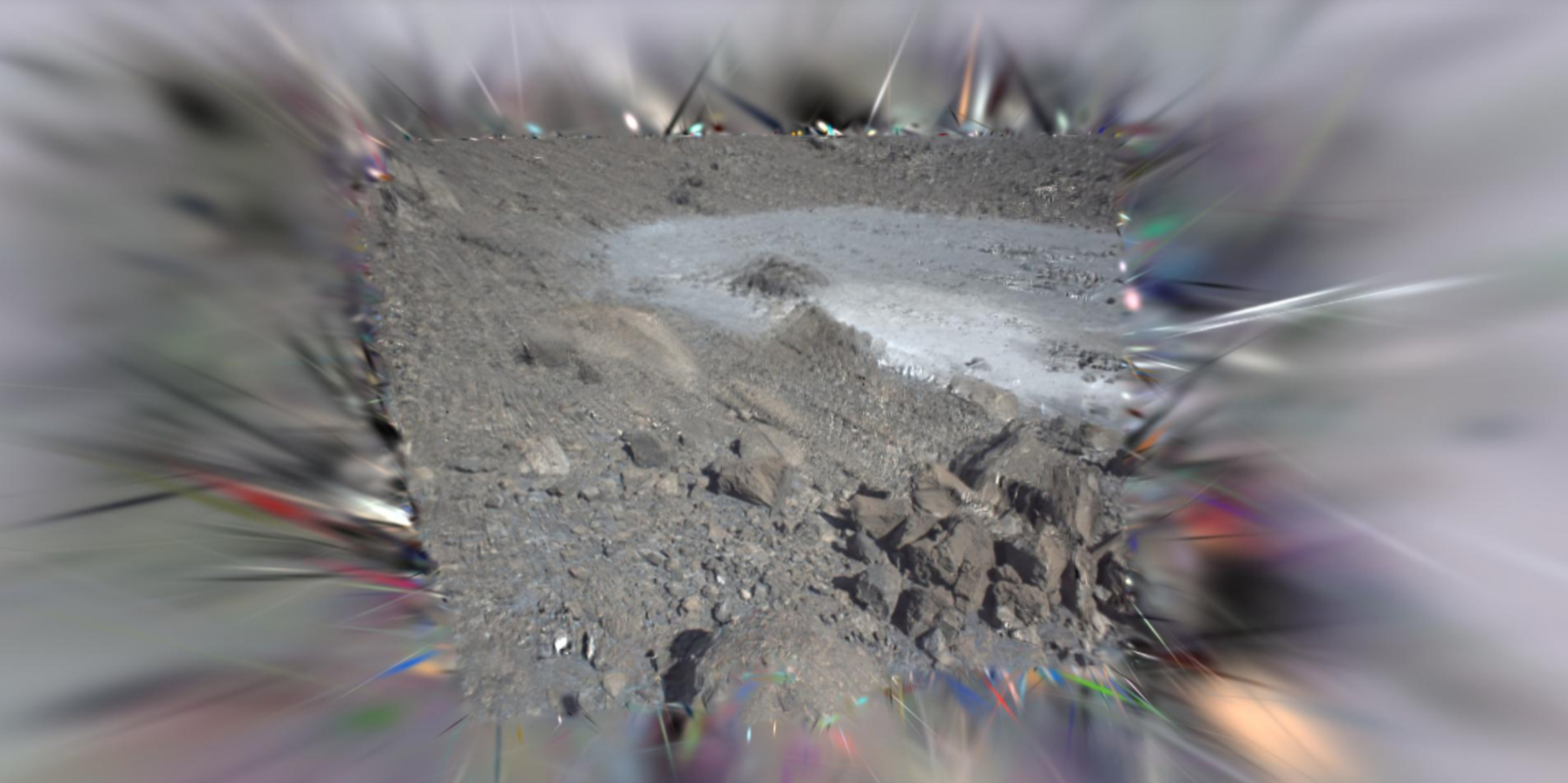} &
\img{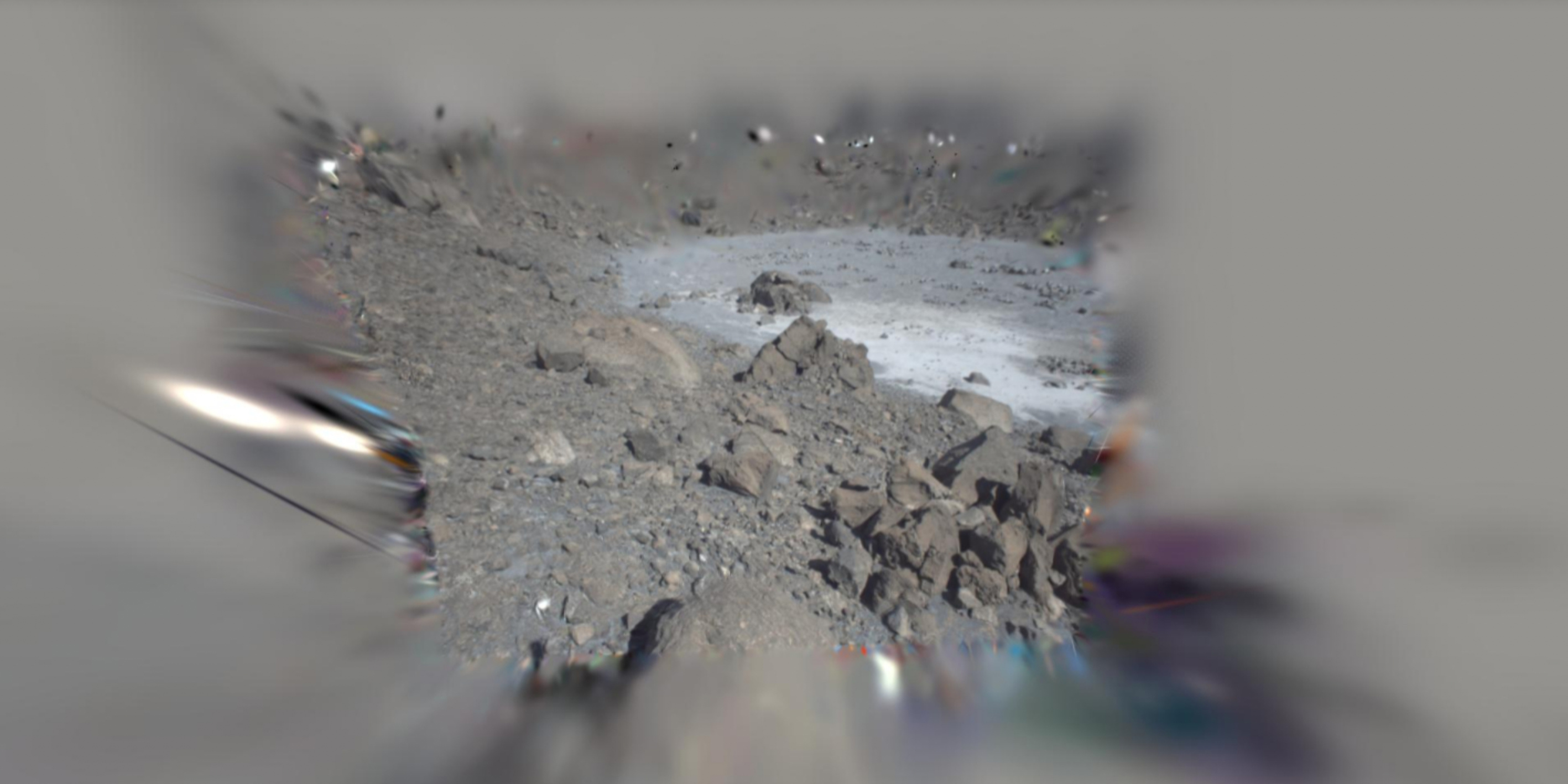} &
\img{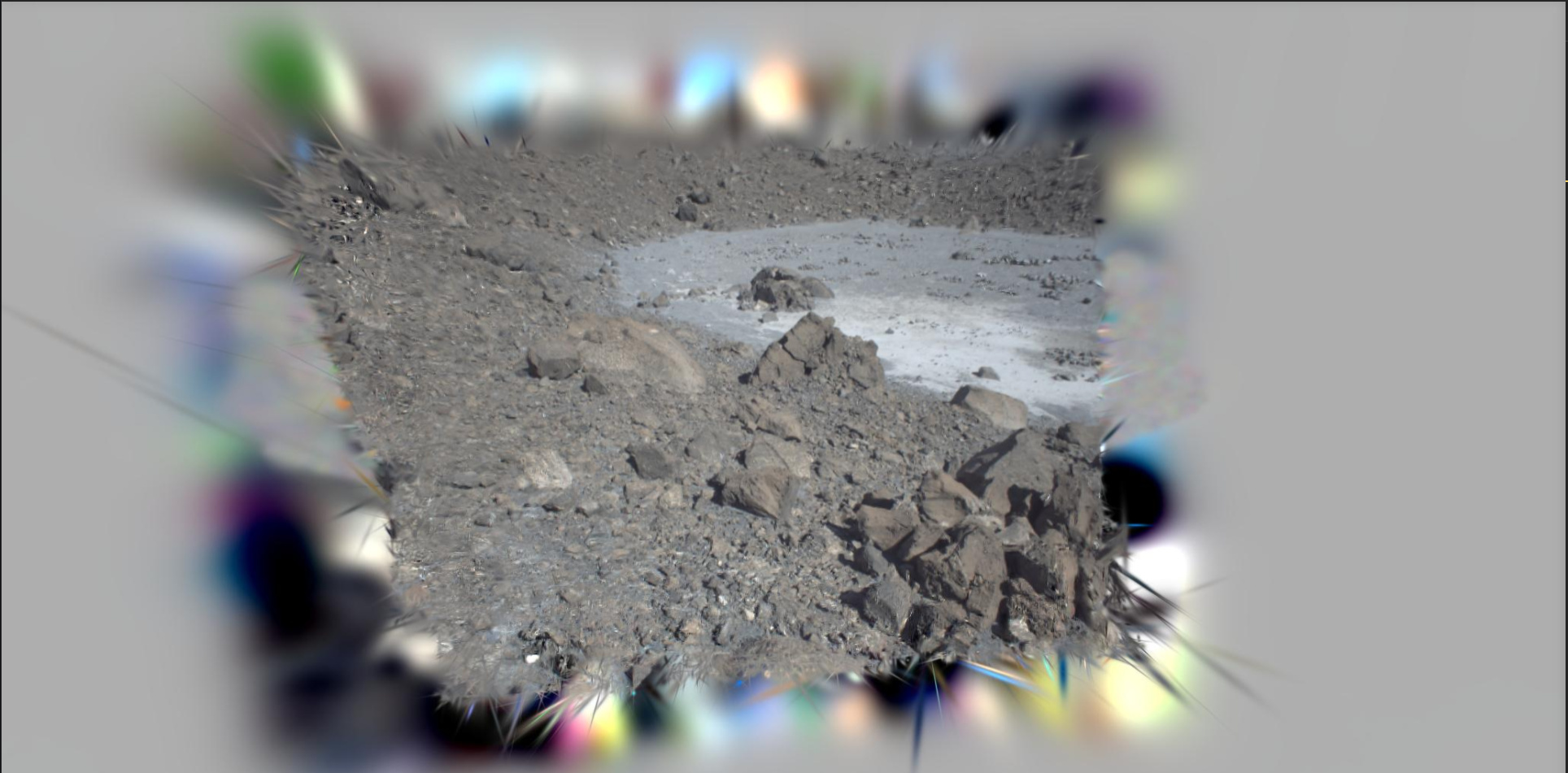} &
\img{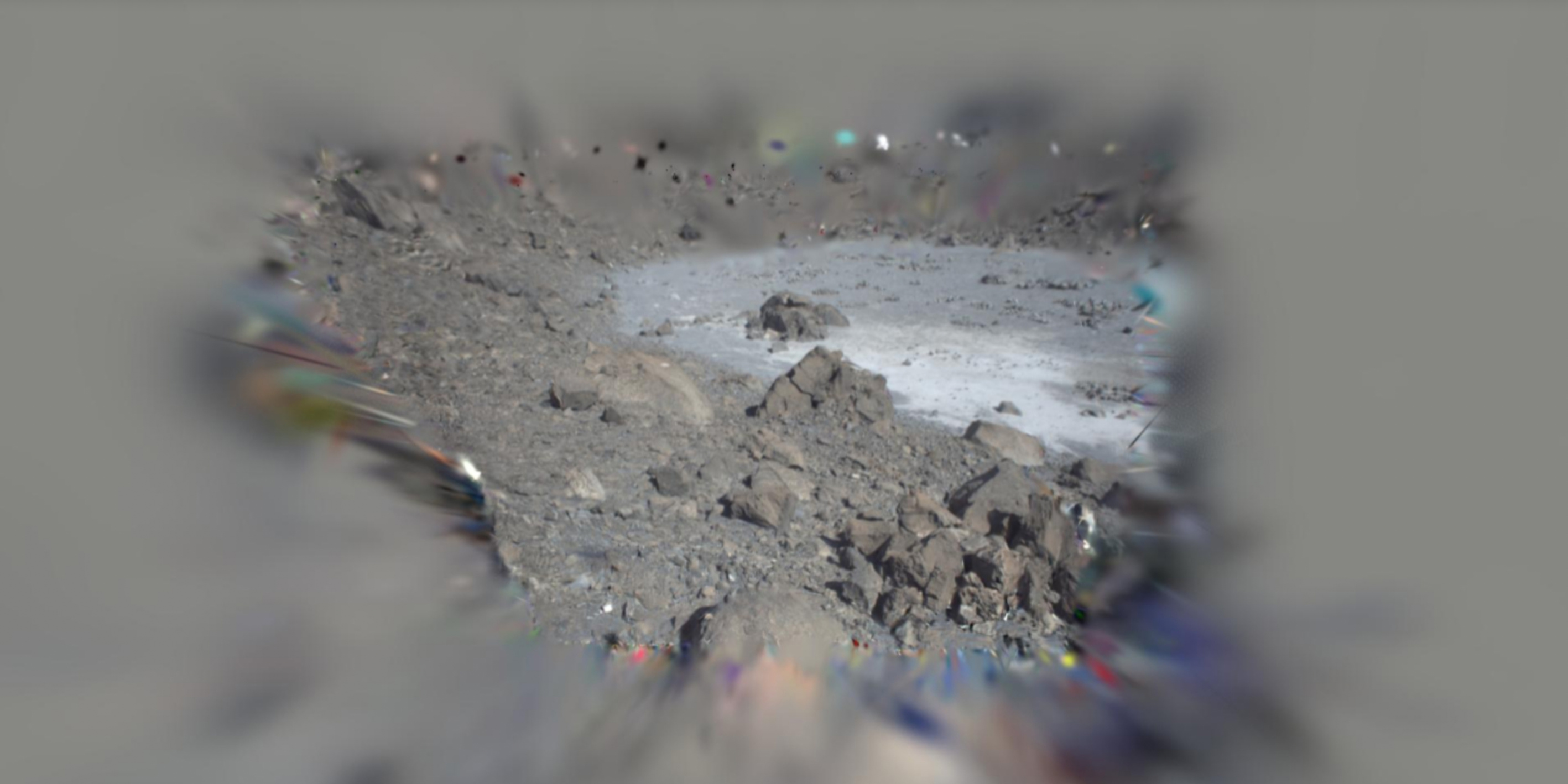} \\[-0.05mm]

\rowlabel{Depth}\rowstrut &
&
\img{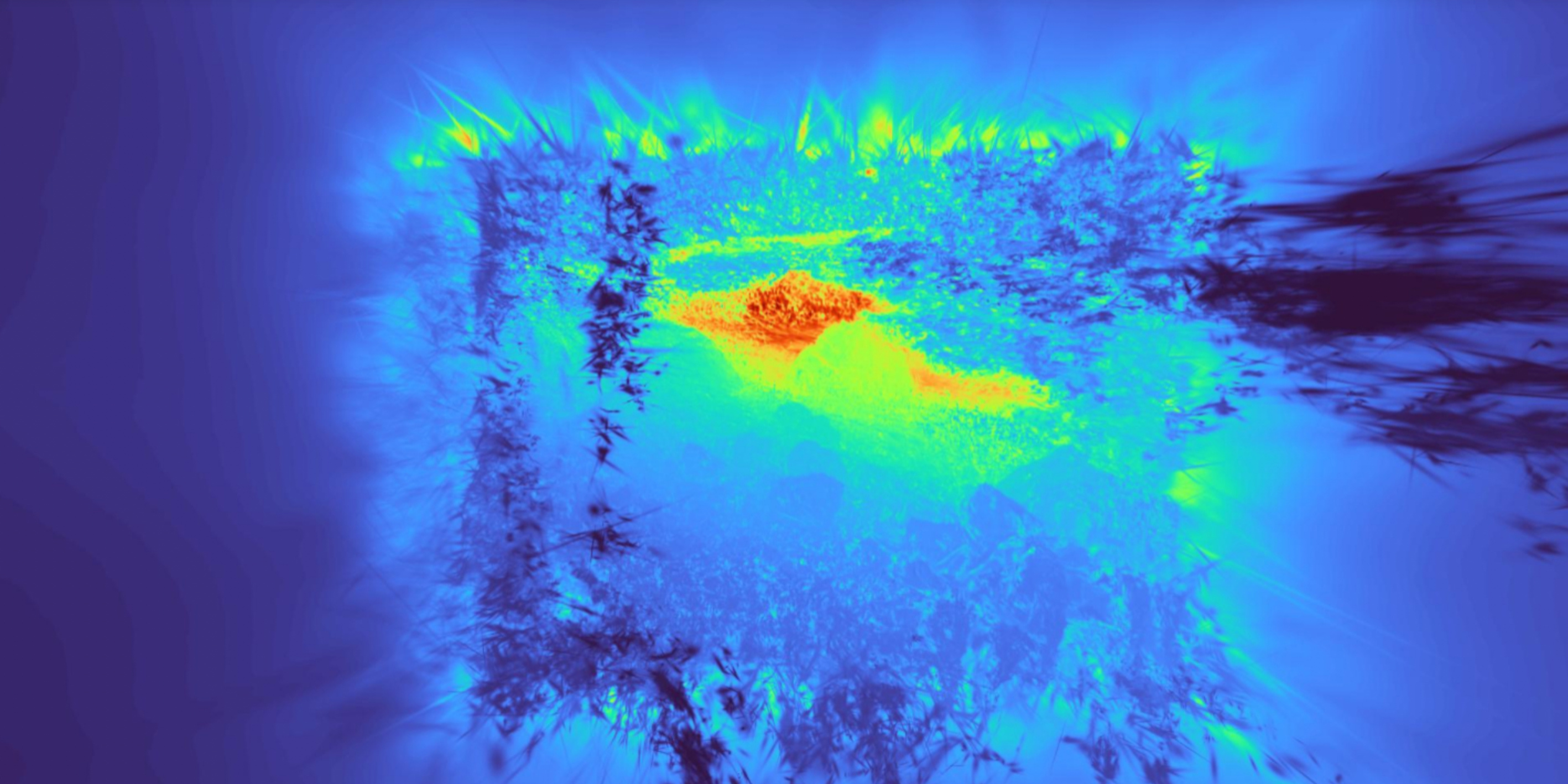} &
\img{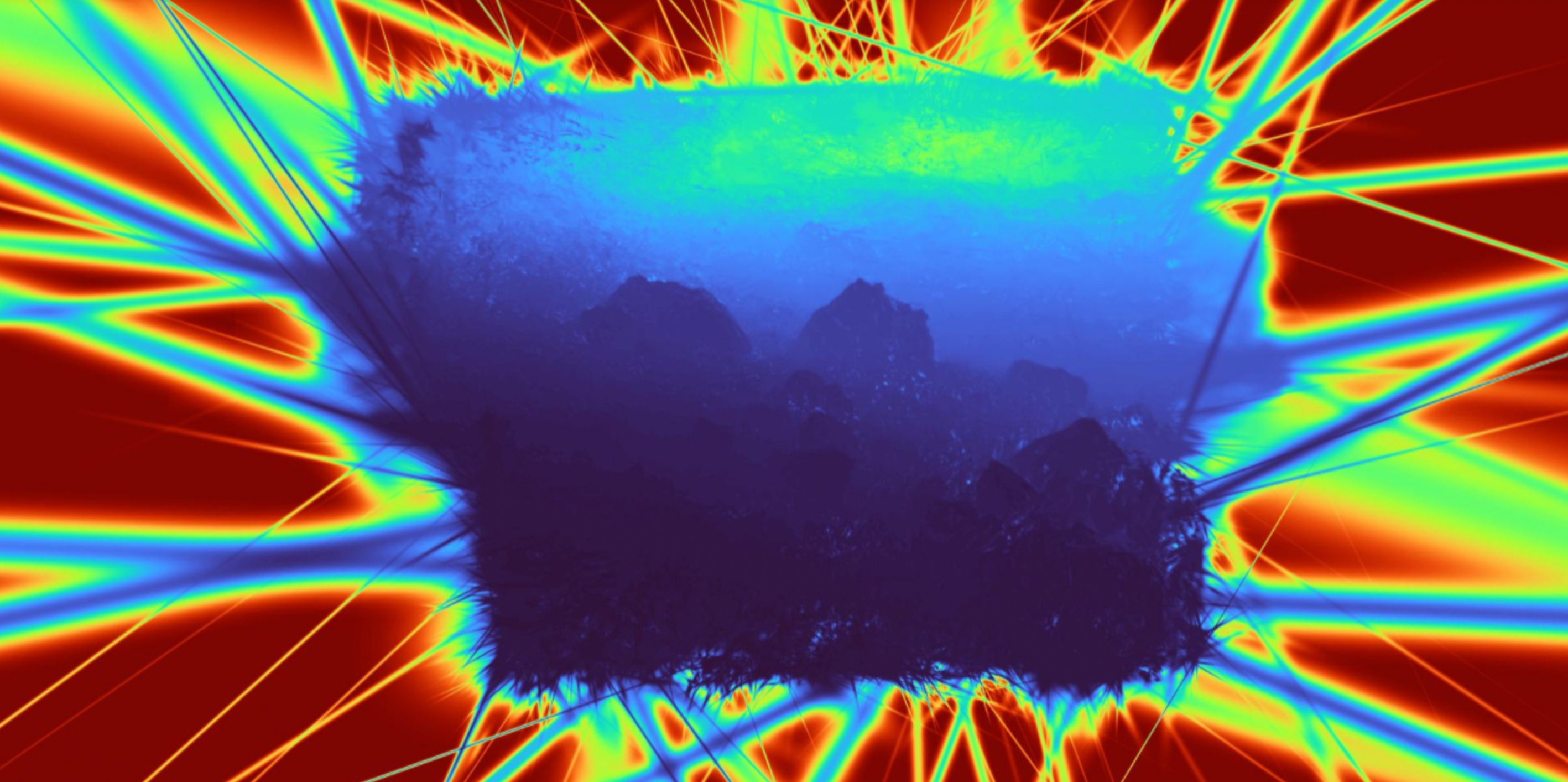} &
\img{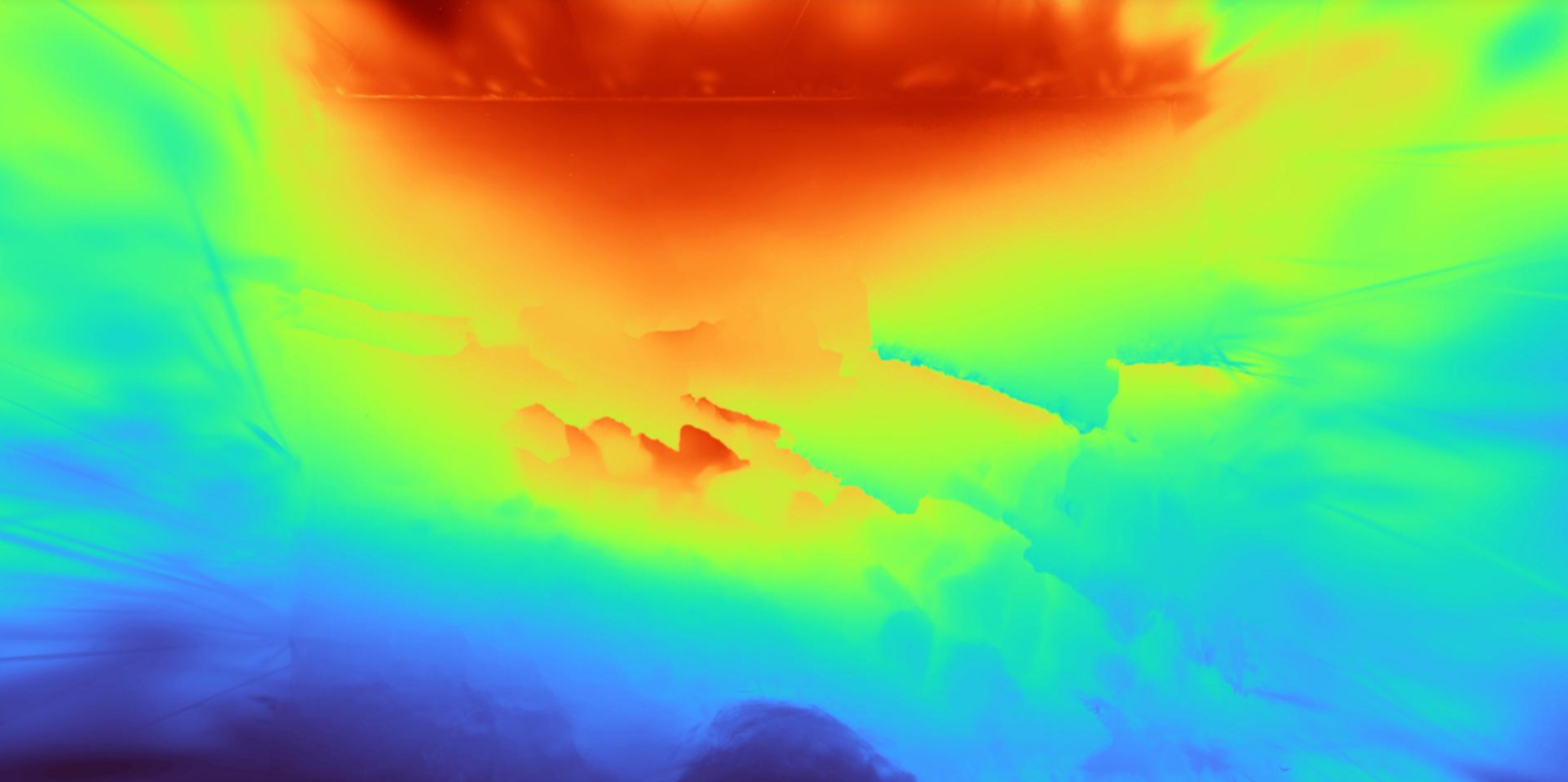} &
\img{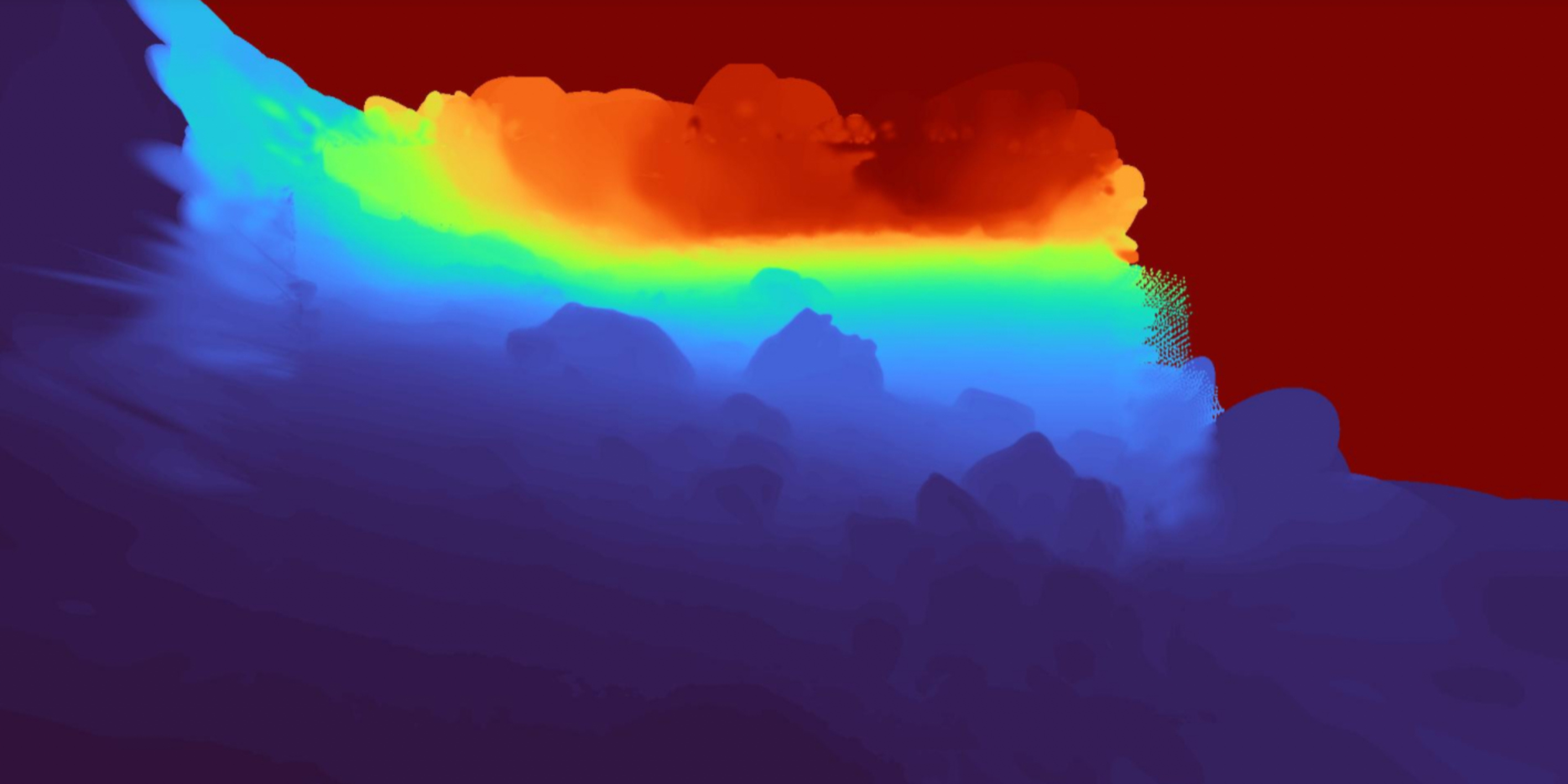} &
\img{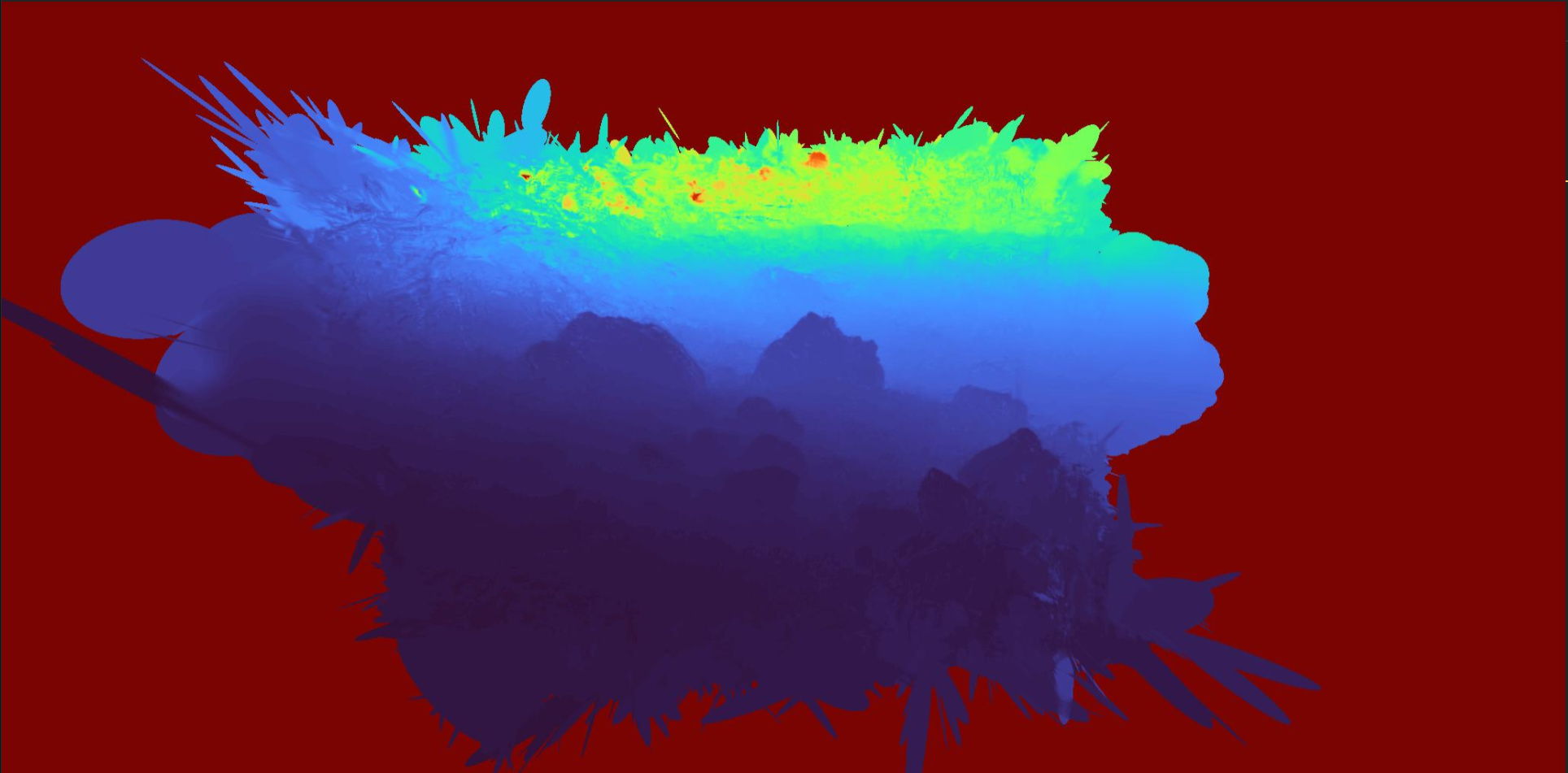} &
\img{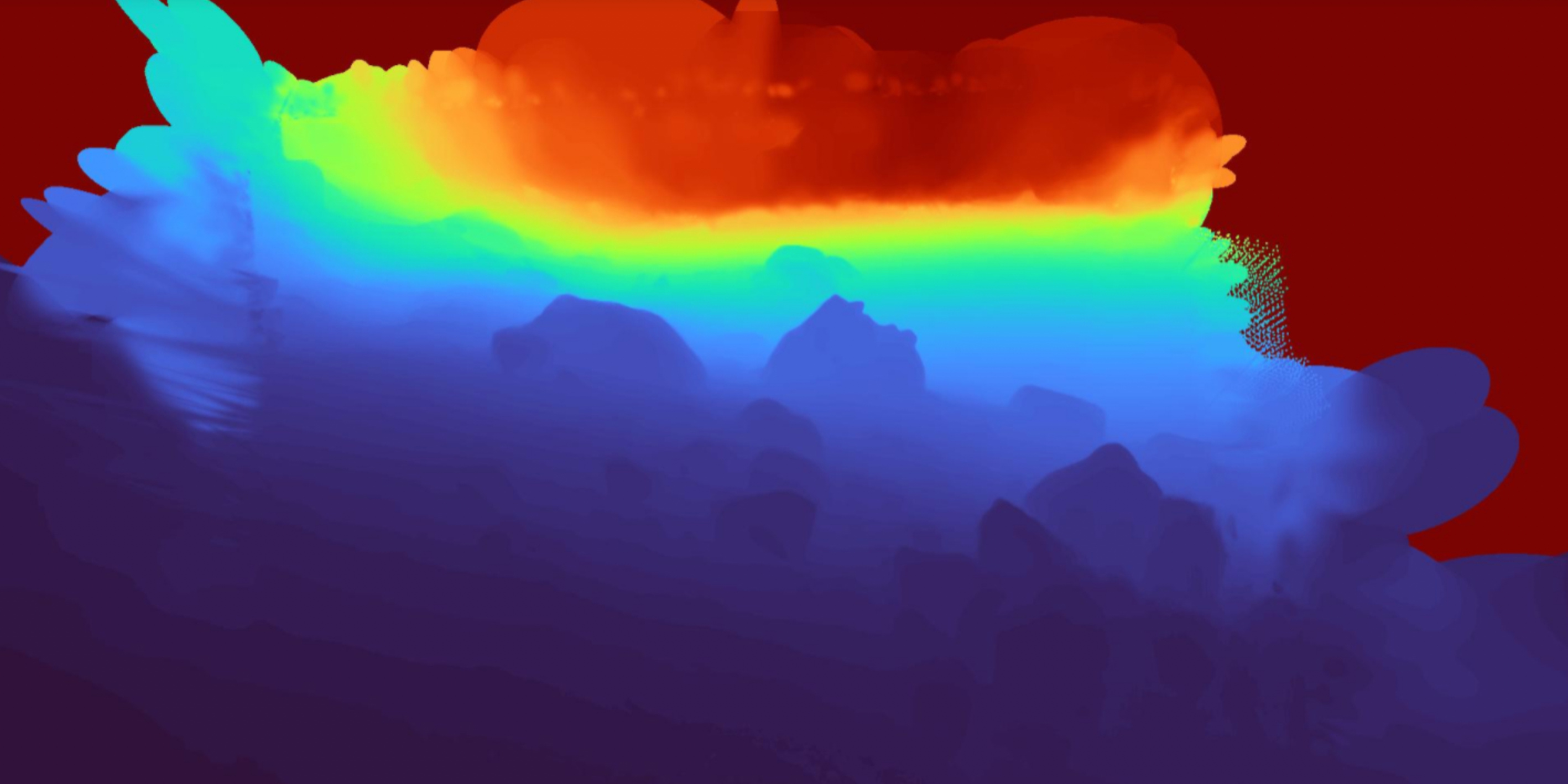} \\

\end{tabular}

\caption{\textbf{Qualitative comparison of Gaussian Splatting variants on two representative submaps.} Rows alternate RGB renderings and depth maps (blue denotes closer regions and red denotes farther ones), while columns compare different supervision strategies. The rectangle overlaid on the RGB rendering of the 3DGS baseline indicates the field of view of the test camera, highlighting the region directly comparable to the ground-truth image. Our method exhibits markedly improved photometric quality and geometric consistency compared to alternatives with weaker geometric supervision.}
\label{fig:qualitative_results}
\end{figure*}

Fig.~\ref{fig:qualitative_results} presents a qualitative comparison of RGB renderings and depth maps obtained with different training variants in representative examples. From left to right, the photometric-only \textbf{3DGS} baseline exhibits blurred surfaces, floating Gaussians, and inconsistent depth structures, particularly in low-texture areas. Initializing the Gaussian centers with a LiDAR prior (\textbf{3DGS + LiDAR}) improves structural coherence; however, noticeable noise and local artifacts persist.
Introducing MVSA supervision (\textbf{3DGS + MVSA}) enhances local surface smoothness and depth consistency, but depth estimates in distant regions remain unstable, leading to visible rendering artifacts at larger ranges. Incorporating the LiDAR prior (\textbf{3DGS + MVSA + LiDAR}) further improves global geometric consistency, especially in far-field areas. Applying the proposed Chamfer loss without MVSA (\textbf{3DGS + Chamfer + LiDAR}) better anchors the reconstruction to the LiDAR geometry, reducing large-scale geometric distortions despite lower local accuracy. Finally, our full method (\textbf{Ours}) combines all components, producing the best geometric accuracy while preserving high photometric fidelity.

\begin{figure}[t]
    \centering
    \includegraphics[width=\linewidth]{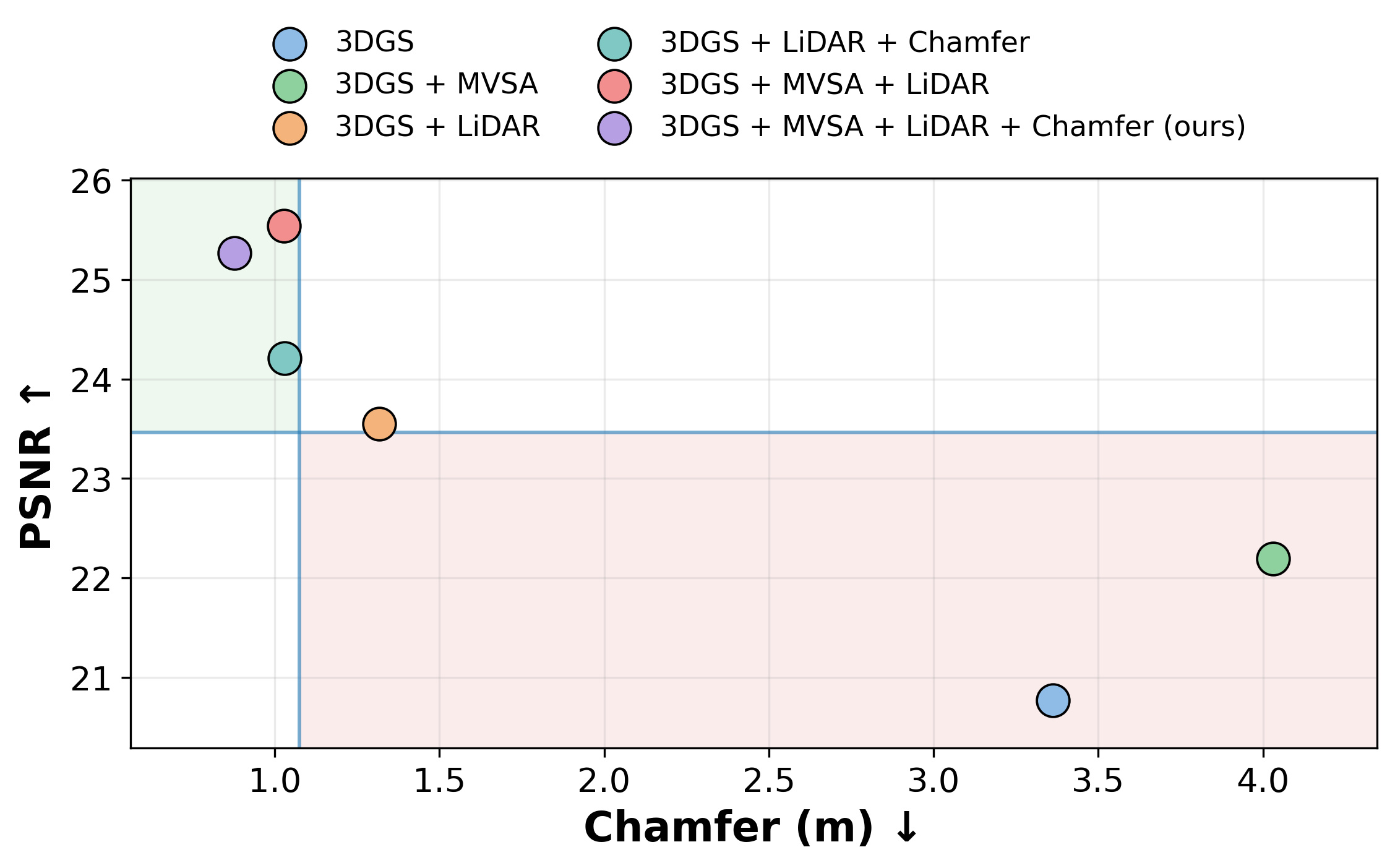}
    \caption{\textbf{Rendering quality (vertical axis) versus geometric accuracy (horizontal axis)} under different photometric and geometric supervision alternatives. The top-left region indicates better performance. As claimed, stronger geometric supervision substantially improves both metrics. Additionally, note that MVSA and LiDAR supervision exhibit complementary effects, and their combination achieves the best overall results.}
    \label{fig:phot_vs_geometric}
\end{figure}

Fig. \ref{fig:phot_vs_geometric} presents an aggregated quantitative analysis of the photometric–geometric trade-off. The baseline 3DGS model achieves the lowest (worst) PSNR, highlighting the limitations of purely photometric supervision. MVSA and LiDAR supervision play complementary roles: MVSA improves photometric reconstruction but degrades geometric accuracy, yielding the worst Chamfer distance among all variants, whereas LiDAR is the primary driver of geometric fidelity, lowering the Chamfer distance while also improving photometric metrics. Their combination (3DGS + MVSA + LiDAR) exploits this complementarity and achieves the best photometric quality. Finally, adding this combination with our LiDAR-guided Chamfer loss (Ours) yields the lowest Chamfer distance overall, at the cost of only a marginal, statistically non-significant decrease in PSNR.

Quantitatively, Table~\ref{tab:chamfer_results} shows that LiDAR-based initialization reduces the geometric error by 61\% with respect to the original 3DGS baseline, while the addition of Chamfer-based supervision lowers the error by roughly 74\%. 
Importantly, these geometric improvements are obtained without compromising photometric fidelity. This confirms that metric consistency, rather than photometric reconstruction quality alone, is a key factor for achieving high-quality representations. Moreover, it shows for the first time that combining MVS and LiDAR supervision yields a more accurate and consistent model than either source of geometric information used independently.

\begin{table}[t]
\caption{Average photometric and geometric metrics across submaps, for different geometric supervision.}
\label{tab:chamfer_results}
\centering
\setlength{\tabcolsep}{3pt}
\begin{tabularx}{\linewidth}{Xcccc}
\toprule
\textbf{Method} & \textbf{Chamfer} $\downarrow$ & \textbf{PSNR} $\uparrow$ & \textbf{SSIM} $\uparrow$ & \textbf{LPIPS}$\downarrow$ \\
\midrule
3DGS & 3.36 & 20.77 & 0.52 & 0.26 \\
3DGS + MVSA & 4.03 & 22.19 & 0.49 & 0.27\\
3DGS + LiDAR & 1.32 & 23.55 & 0.67  & \underline{0.18}\\
3DGS + Chamfer + LiDAR & 1.03 & 24.21 & 0.70 & \textbf{0.17}\\
3DGS + MVSA + LiDAR & \underline{1.02} & \textbf{25.54} & \textbf{0.74} & \textbf{0.17}\\
3DGS + MVSA + LiDAR + Chamfer (ours) & \textbf{0.88} & \underline{25.27} & \underline{0.72} & \underline{0.18}\\
\bottomrule
\multicolumn{5}{l}{\footnotesize Best results are \textbf{bolded}, second best are \underline{underlined}.}\\
\end{tabularx}
\end{table}

\subsection{Relocalization Results}

We now evaluate how geometric and photometric consistency impacts relocalization accuracy. 
We compare our geometry-aware 3DGS representation against the photometric-only baseline, while keeping the place recognition pipeline and the 6DGS pose estimator fixed. Therefore, performance differences directly reflect the impact of geometric consistency in the underlying map representation.

\begin{figure}[t!]
    \centering
    \includegraphics[width=\linewidth]{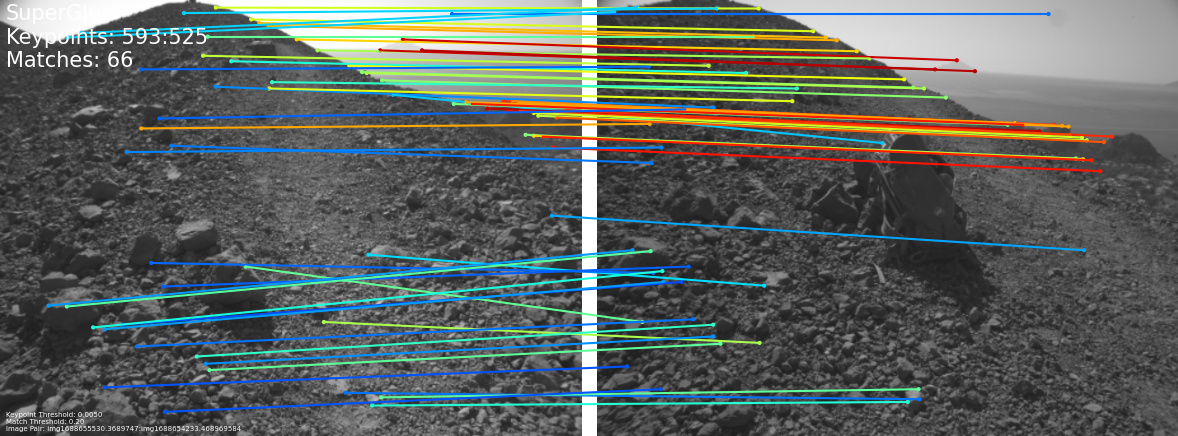}
    \caption{\textbf{Illustrative example of SuperPoint and SuperGlue feature matching between a query and a retrieved image.} Several incorrect correspondences are observed, particularly at close range, caused by aliasing and significant viewpoint changes. }
    \label{fig:superglue_matches}
\end{figure}

We additionally evaluate a classical relocalization pipeline based on feature matching followed by PnP. Fig.~\ref{fig:superglue_matches} illustrates a representative matching example. Despite the abundance of texture, resulting in large amounts of SuperPoint features, only a few reliable correspondences are established by SuperGlue. As a result, the geometric constraints are weak or outlier-contaminated, and the overall relocalization performance degrades significantly.

Table~\ref{tab:pose_recall_thresholds} reports pose recall. 
Under the relaxed (10m, 15°) threshold, both PnP, 3DGS and 3DGS with LiDAR prior exhibit extremely limited recall (0.16\%, 6.25\%, 2.10\% respectively), whereas the proposed method reaches 43.2\%. 
This indicates that the geometry-aware map enables the pose optimizer to successfully refine moderately misaligned retrieval candidates.
Under the stricter (2m, 10°) threshold, both baselines completely fail (0.0\% recall), while our proposed methods, with stronger geometric supervision, yield higher recall ranges, with our method achieving the best result (6.80\%).

\begin{table}[t]
\centering
\caption{Pose recall comparison between baselines PnP and 3DGS, and our geometry-aware 3DGS model.}
\setlength{\tabcolsep}{4pt}
\begin{tabularx}{\linewidth}{Xcc}
\toprule
\textbf{Method} & \textbf{Recall (10m, 15°)} & \textbf{Recall (2m, 10°)} \\
\midrule
   PnP     & 0.16 & 0.00 \\
 3DGS & 6.25 & 0.00 \\
 3DGS + LiDAR & 2.10 & 0.00 \\
 3DGS + MVSA & 22.90 & 6.20 \\
 3DGS + MVSA + LiDAR & 31.20 & 6.20 \\
 3DGS + Chamfer + LiDAR & 28.26 & 6.52 \\
 Ours     & \textbf{43.20} & \textbf{6.80} \\
\bottomrule
\end{tabularx}

\label{tab:pose_recall_thresholds}
\end{table}

\begin{figure}[t]
    \centering
    \includegraphics[width=\linewidth, trim=0mm 33mm 0mm 0mm, clip]{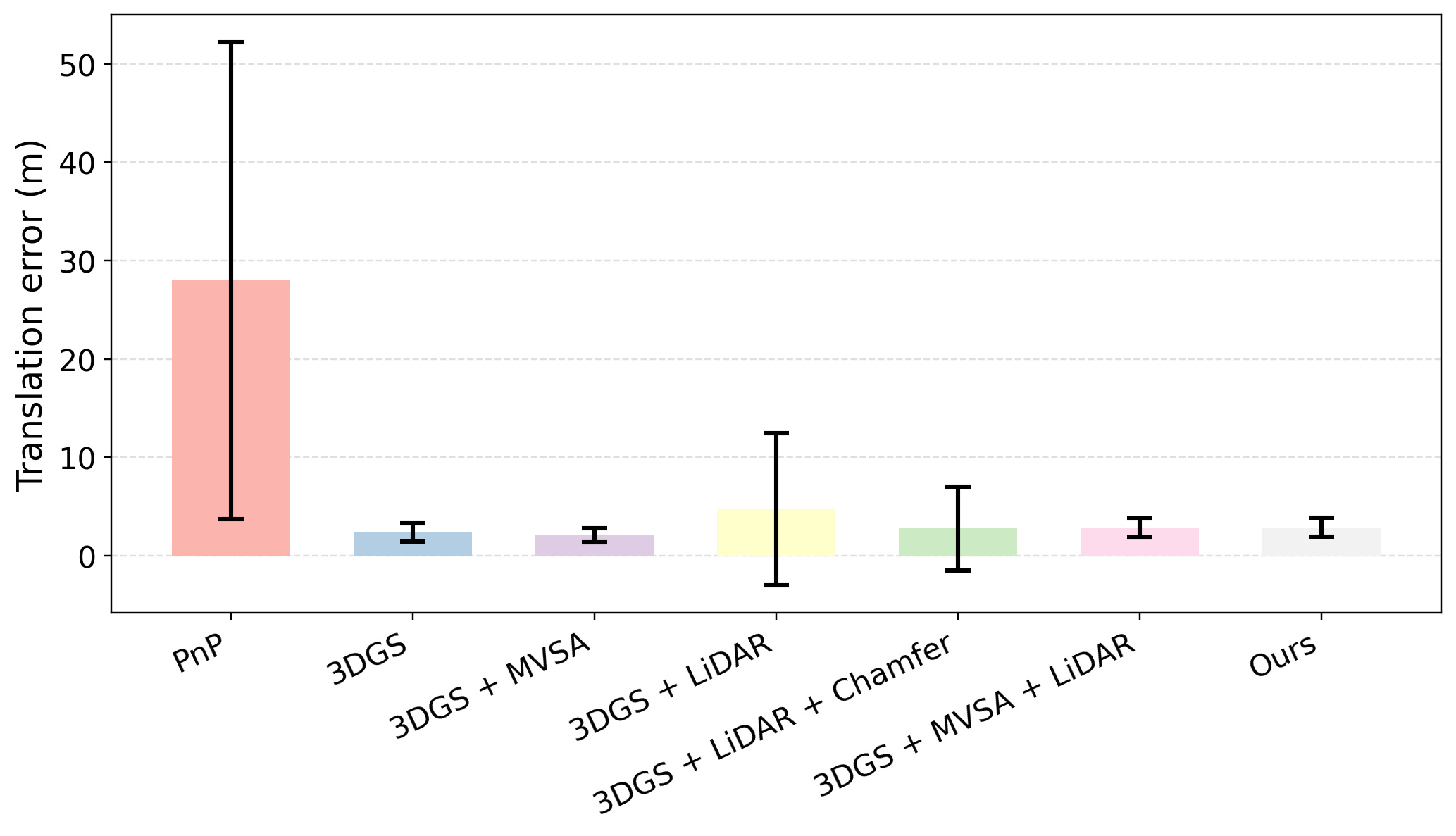}\\ 
    \includegraphics[width=\linewidth, clip]{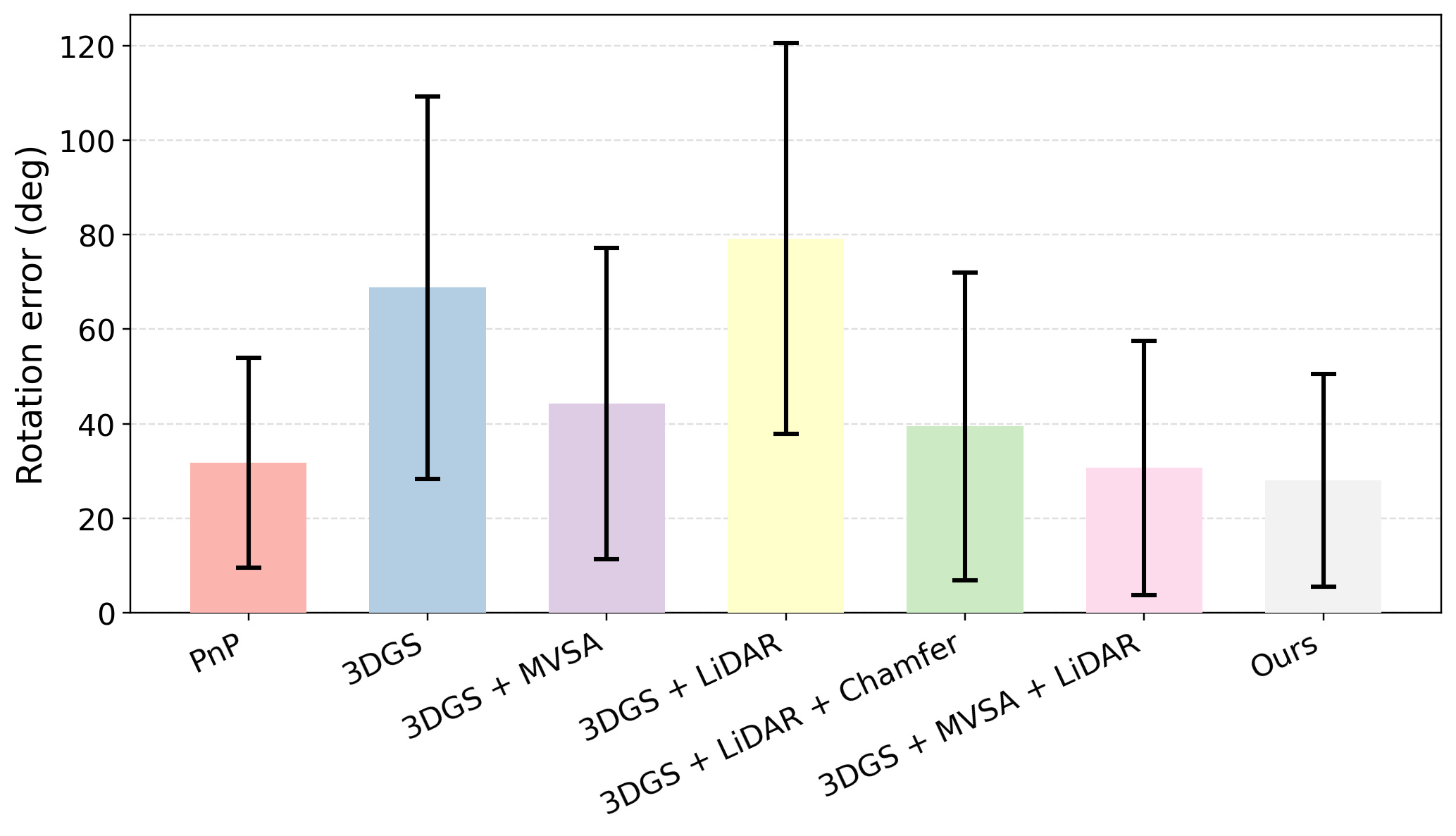} \\

    \caption{\textbf{Translation and rotation estimation error distribution} for relocalization baselines and ours.}
    \label{tab:pose_errors}
\end{figure}

Fig.~\ref{tab:pose_errors} further details the pose errors for each of the methods. 
PnP exhibits large translation errors, while maintaining stable yaw errors (around 30º). This indicates that the lack of reliable depth correspondences, due to the small number of matched keypoints, impacts mainly position estimates.
In contrast, the baseline 3DGS model exhibits consistently large rotation errors, while conserving lower translational errors across all ranks. 
This behavior indicates severe geometric inconsistencies in the photometric-only reconstruction, where visually similar candidates frequently correspond to incorrect spatial alignments.
The proposed geometry-aware representation substantially reduces rotation errors at all ranks, with median values consistently below $25^\circ$, while maintaining competitive translation accuracy. 
This demonstrates that enforcing global metric consistency in the Gaussian map is critical for stable and accurate 6-DoF pose recovery.

\begin{figure}[t]
    \centering
    \begin{minipage}{\linewidth}
    \centering
        \footnotesize 
        \begin{tabular}{p{0.28\linewidth} p{0.28\linewidth} p{0.32\linewidth}}
            \textbf{Query Image} & 
            \textbf{Render (GT Pose)} & 
            \textbf{Render (Est. Pose)}
        \end{tabular}
    \end{minipage}
    \includegraphics[width=\linewidth, trim=10mm 25mm 10mm 35mm, clip]{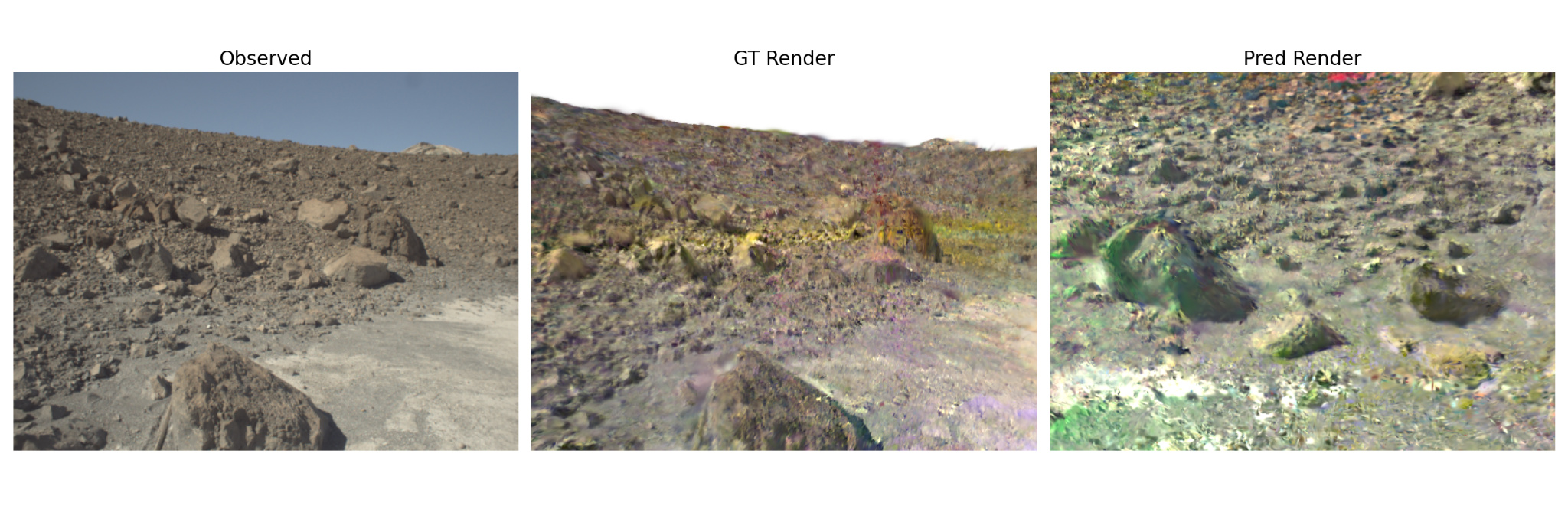}\\ 
    \includegraphics[width=\linewidth, trim=10mm 25mm 10mm 35mm, clip]{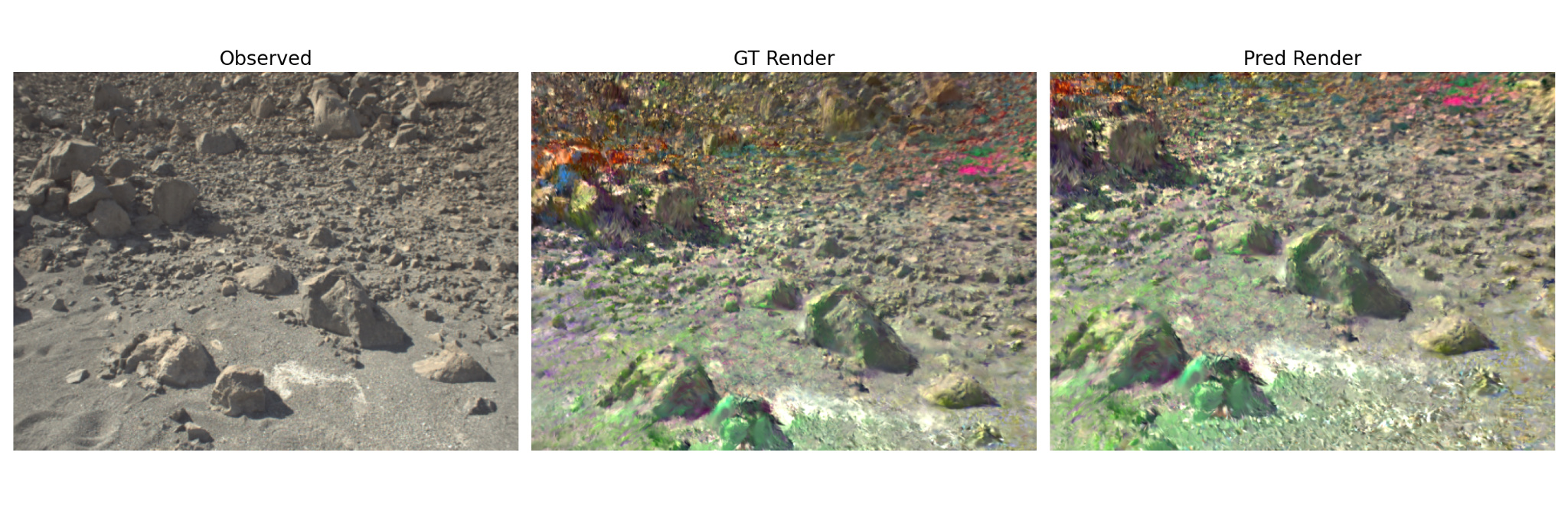} \\
    \includegraphics[width=\linewidth, trim=10mm 25mm 10mm 35mm, clip]{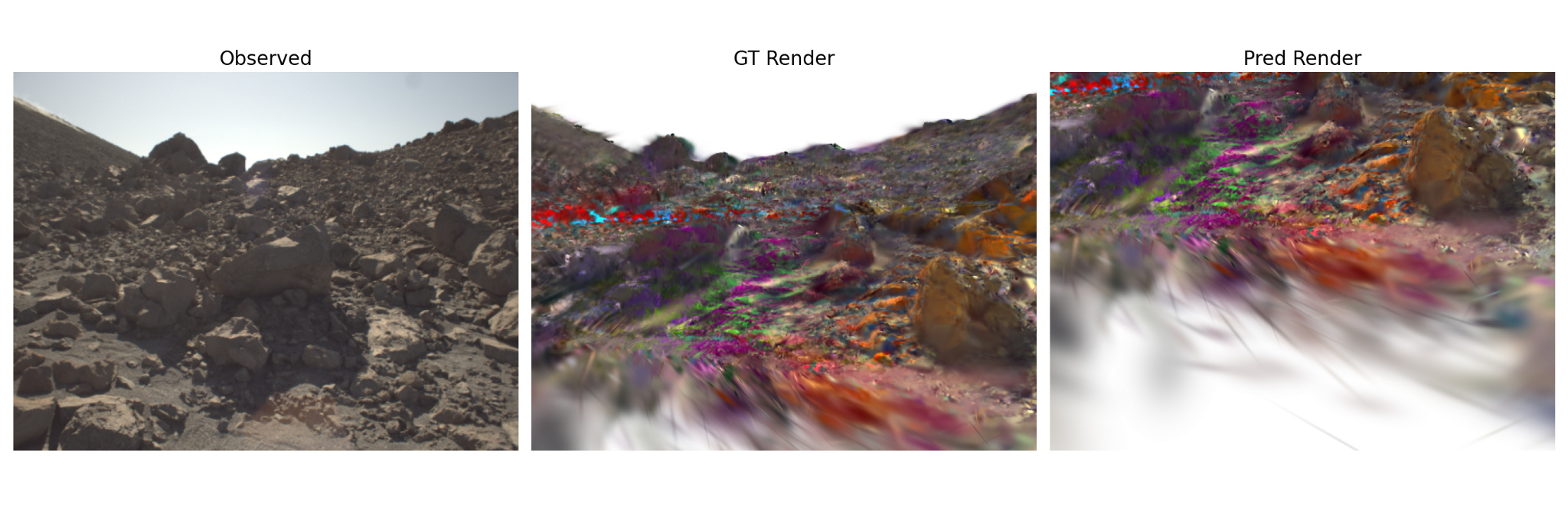}\\
    \includegraphics[width=\linewidth, trim=10mm 25mm 10mm 35mm, clip]{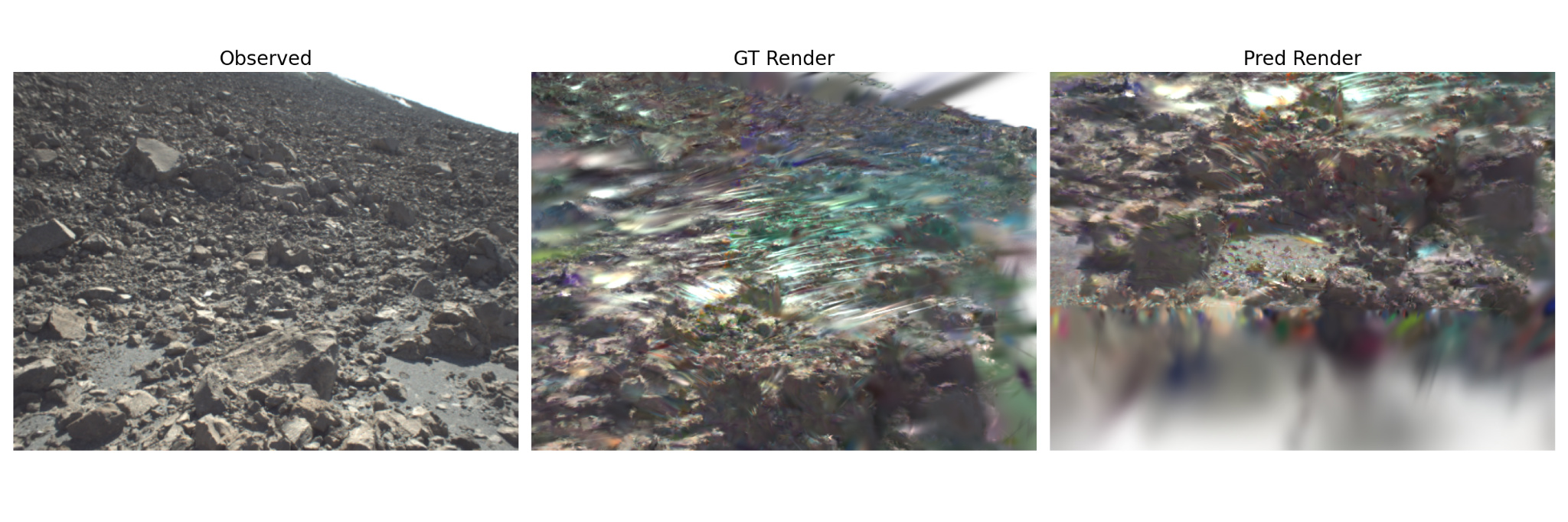}
    \caption{\textbf{Qualitative relocalization examples.} Each row corresponds to a different query. From left to right within each image: observed image,
    ground-truth render, and render obtained from the estimated pose.}
    \label{fig:qual_reloc_examples}
\end{figure}

Fig.~\ref{fig:qual_reloc_examples} presents qualitative relocalization examples. Even when rendering from the ground-truth pose, the synthesized views do not perfectly match the corresponding query images. 
This discrepancy is likely due to local inaccuracies in the 3DGS models, close to the submap boundaries, and the low parallax between viewpoints during training. These examples, together with the low recall rates in Table~\ref{tab:pose_recall_thresholds}, evidence the challenge of the data.
 
Furthermore, the rendered views may exhibit photometric artifacts. This is expected since 6DGS does not refine camera poses via photometric alignment, but rather relying on sparse geometric ray correspondences derived from the underlying Gaussian representation.
Even when appearance fidelity is imperfect, the preserved geometric structure of the scene remains sufficient to recover stable 6-DoF camera poses.

The examples include both successful and failed estimations. Interestingly, even in failure cases, the reconstructed geometry often remains partially aligned with the scene structure, suggesting that residual errors are primarily due to local ambiguities or insufficient geometric constraints rather than catastrophic map degradation.

\section{CONCLUSIONS}

We introduced a relocalization approach tailored to challenging outdoor, planetary-like environments, where low-texture surfaces, perceptual aliasing, and sparse viewpoints severely degrade existing pipelines. By leveraging 3D Gaussian Splatting as an explicit and differentiable scene representation, we show that purely photometric supervision leads to geometrically inconsistent and metrically unreliable reconstructions. In contrast, our geometry-aware training framework, which combines depth and surface normal supervision with a LiDAR-guided Chamfer alignment, enforces structural fidelity and global scale consistency.

Experiments on a real-world planetary rover dataset show that enforcing geometric consistency reduces reconstruction errors over $74\%$ and improves pose recall from $6.25\%$ to $43.20\%$ under challenging conditions. Our findings highlight that geometric consistency, rather than photometric fidelity alone, is the key factor for reliable 6-DoF relocalization in extreme outdoor environments. 







\balance
\bibliographystyle{IEEEtran}
\bibliography{references}

@article{lepetit2009epnp,
author = {Lepetit, Vincent and Moreno-Noguer, Francesc and Fua, Pascal},
title = {EPnP: An Accurate O(n) Solution to the PnP Problem},
journal = {International Journal of Computer Vision},
volume = {81},
number = {2},
pages = {155--166},
year = {2009},
doi = {10.1007/s11263-008-0152-6}
}

@article{fischler1981random,
author = {Fischler, Martin A. and Bolles, Robert C.},
title = {Random sample consensus: a paradigm for model fitting with applications to image analysis and automated cartography},
year = {1981},
volume = {24},
number = {6},
journal = {Commun. ACM},
pages = {381–395},
numpages = {15},
}

@article{kerbl20233dgaussiansplattingrealtime,
  title={3d gaussian splatting for real-time radiance field rendering.},
  author={Kerbl, Bernhard and Kopanas, Georgios and Leimk{\"u}hler, Thomas and Drettakis, George and others},
  journal={ACM Trans. Graph.},
  volume={42},
  number={4},
  pages={139--1},
  year={2023}
}

@ARTICLE{DLRdataset,
  author={Giubilato, Riccardo and Stürzl, Wolfgang and Wedler, Armin and Triebel, Rudolph},  
  journal={IEEE Robotics and Automation Letters},   
  title={Challenges of SLAM in extremely unstructured environments: the DLR Planetary Stereo, Solid-State LiDAR, Inertial Dataset},   
  year={2022},  
  volume={},  
  number={},  
  pages={1-8},  
  doi={10.1109/LRA.2022.3188118}}

@article{kendall2015posenet,
  author       = {Alex Kendall and
                  Matthew Grimes and
                  Roberto Cipolla},
  title        = {Convolutional networks for real-time 6-DOF camera relocalization},
  journal      = {CoRR},
  volume       = {abs/1505.07427},
  year         = {2015},
  eprinttype    = {arXiv},
  eprint       = {1505.07427},
  bibsource    = {dblp computer science bibliography, https://dblp.org}
}

@article{tewari2022advances,
author = {Tewari, A. and Thies, J. and Mildenhall, B. and Srinivasan, P. and Tretschk, E. and Yifan, W. and Lassner, C. and Sitzmann, V. and Martin-Brualla, R. and Lombardi, S. and Simon, T. and Theobalt, C. and Nießner, M. and Barron, J. T. and Wetzstein, G. and Zollhöfer, M. and Golyanik, V.},
title = {Advances in Neural Rendering},
journal = {Computer Graphics Forum},
volume = {41},
number = {2},
pages = {703-735},
year = {2022}
}

@inproceedings{chen1995view,
author = {Chen, Shenchang Eric and Williams, Lance},
title = {View interpolation for image synthesis},
year = {1993},
booktitle = {Proceedings of the 20th Annual Conference on Computer Graphics and Interactive Techniques},
pages = {279–288},
numpages = {10},
series = {SIGGRAPH '93}
}

@inproceedings{seitz1996view,
author = {Seitz, Steven M. and Dyer, Charles R.},
title = {View morphing},
year = {1996},
booktitle = {Proceedings of the 23rd Annual Conference on Computer Graphics and Interactive Techniques},
pages = {21–30},
numpages = {10},
series = {SIGGRAPH '96}
}

@inproceedings{yen2020inerf,
  title={{iNeRF}: Inverting Neural Radiance Fields for Pose Estimation},
  author={Lin Yen-Chen and Pete Florence and Jonathan T. Barron and Alberto Rodriguez and Phillip Isola and Tsung-Yi Lin},
  booktitle={IEEE/RSJ International Conference on Intelligent Robots and Systems ({IROS})},
  year={2021}
}

@article{NeRF,
  title={Nerf: Representing scenes as neural radiance fields for view synthesis},
  author={Mildenhall, Ben and Srinivasan, Pratul P and Tancik, Matthew and Barron, Jonathan T and Ramamoorthi, Ravi and Ng, Ren},
  journal={Communications of the ACM},
  volume={65},
  number={1},
  pages={99--106},
  year={2021},
  publisher={ACM New York, NY, USA}
}

@misc{jiang2025gslidargeneratingrealisticlidar,
      title={GS-LiDAR: Generating Realistic LiDAR Point Clouds with Panoramic Gaussian Splatting}, 
      author={Junzhe Jiang and Chun Gu and Yurui Chen and Li Zhang},
      year={2025},
      eprint={2501.13971},
      archivePrefix={arXiv},
      primaryClass={cs.CV},
}

@inproceedings{hess2025splatadrealtimelidarcamera,
  title={SplatAD: Real-time lidar and camera rendering with 3d gaussian splatting for autonomous driving},
  author={Hess, Georg and Lindstr{\"o}m, Carl and Fatemi, Maryam and Petersson, Christoffer and Svensson, Lennart},
  booktitle={Proceedings of the Computer Vision and Pattern Recognition Conference},
  pages={11982--11992},
  year={2025}
}

@inproceedings{izquierdo2025mvsanywhere,
  title={Mvsanywhere: Zero-shot multi-view stereo},
  author={Izquierdo, Sergio and Sayed, Mohamed and Firman, Michael and Garcia-Hernando, Guillermo and Turmukhambetov, Daniyar and Civera, Javier and Mac Aodha, Oisin and Brostow, Gabriel and Watson, Jamie},
  booktitle={Proceedings of the Computer Vision and Pattern Recognition Conference},
  pages={11493--11504},
  year={2025}
}

@inproceedings{izquierdo2024optimaltransportaggregationvisual,
  title={Optimal transport aggregation for visual place recognition},
  author={Izquierdo, Sergio and Civera, Javier},
  booktitle={Proceedings of the ieee/cvf conference on computer vision and pattern recognition},
  pages={17658--17668},
  year={2024}
}

@article{alejandra2025multi,
  title={Multi-modal Loop Closure Detection with Foundation Models in Severely Unstructured Environments},
  author={Gonzalez, Laura Alejandra Encinar and Folkesson, John and Triebel, Rudolph and Giubilato, Riccardo},
  journal={2026 IEEE international conference on robotics and automation (ICRA)},
  year={2026}
}

@misc{bortolon20246dgs6dposeestimation,
  title={6dgs: 6d pose estimation from a single image and a 3d gaussian splatting model},
  author={Matteo, Bortolon and Tsesmelis, Theodore and James, Stuart and Poiesi, Fabio and Del Bue, Alessio},
  booktitle={European Conference on Computer Vision},
  pages={420--436},
  year={2024},
  organization={Springer}
}

@article{ebadi2023present,
  title={Present and future of SLAM in extreme environments: The DARPA SubT challenge},
  author={Ebadi, Kamak and Bernreiter, Lukas and Biggie, Harel and Catt, Gavin and Chang, Yun and Chatterjee, Arghya and Denniston, Christopher E and Desch{\^e}nes, Simon-Pierre and Harlow, Kyle and Khattak, Shehryar and others},
  journal={IEEE Transactions on Robotics},
  volume={40},
  pages={936--959},
  year={2023},
  publisher={IEEE}
}

@article{geromichalos2020slam,
  title={SLAM for autonomous planetary rovers with global localization},
  author={Geromichalos, Dimitrios and Azkarate, Martin and Tsardoulias, Emmanouil and Gerdes, Levin and Petrou, Loukas and Perez Del Pulgar, Carlos},
  journal={Journal of Field Robotics},
  volume={37},
  number={5},
  pages={830--847},
  year={2020},
  publisher={Wiley Online Library}
}

@article{tosi2026nerfs,
  title={How nerfs and 3d gaussian splatting are reshaping slam: a survey},
  author={Tosi, Fabio and Zhang, Youmin and Gong, Ziren and Mattoccia, Stefano and Oswald, Martin R and Sandstrom, Erik and Poggi, Matteo},
  journal={IEEE Transactions on Robotics},
  year={2026},
  publisher={IEEE}
}

@article{deng2025best,
  title={What is the best 3d scene representation for robotics? from geometric to foundation models},
  author={Deng, Tianchen and Pan, Yue and Yuan, Shenghai and Li, Dong and Wang, Chen and Li, Mingrui and Chen, Long and Xie, Lihua and Wang, Danwei and Wang, Jingchuan and others},
  journal={arXiv preprint arXiv:2512.03422},
  year={2025}
}

@inproceedings{sarlin2020supergluelearningfeaturematching,
  title={Superglue: Learning feature matching with graph neural networks},
  author={Sarlin, Paul-Edouard and DeTone, Daniel and Malisiewicz, Tomasz and Rabinovich, Andrew},
  booktitle={Proceedings of the IEEE/CVF conference on computer vision and pattern recognition},
  pages={4938--4947},
  year={2020}
}

@article{giubilato2026s3li,
  title={The S3LI Vulcano Dataset: A Dataset for Multi-Modal SLAM in Unstructured Planetary Environments},
  author={Giubilato, Riccardo and M{\"u}ller, Marcus Gerhard and Sewtz, Marco and Gonzalez, Laura Alejandra Encinar and Folkesson, John and Triebel, Rudolph},
  journal={2026 IEEE Aerospace Conference},
  year={2026}
}

@inproceedings{giubilato2021multi,
  title={Multi-modal loop closing in unstructured planetary environments with visually enriched submaps},
  author={Giubilato, Riccardo and Vayugundla, Mallikarjuna and St{\"u}rzl, Wolfgang and Schuster, Martin J and Wedler, Armin and Triebel, Rudolph},
  booktitle={2021 IEEE/RSJ International Conference on Intelligent Robots and Systems (IROS)},
  pages={8758--8765},
  year={2021},
  organization={IEEE}
}

@article{douze2025faiss,
  title={The faiss library},
  author={Douze, Matthijs and Guzhva, Alexandr and Deng, Chengqi and Johnson, Jeff and Szilvasy, Gergely and Mazar{\'e}, Pierre-Emmanuel and Lomeli, Maria and Hosseini, Lucas and J{\'e}gou, Herv{\'e}},
  journal={IEEE Transactions on Big Data},
  year={2025},
  publisher={IEEE}
}

@inproceedings{matsuki2024gaussiansplattingslam,
  title={Gaussian splatting slam},
  author={Matsuki, Hidenobu and Murai, Riku and Kelly, Paul HJ and Davison, Andrew J},
  booktitle={Proceedings of the IEEE/CVF conference on computer vision and pattern recognition},
  pages={18039--18048},
  year={2024}
}

@inproceedings{zhu2025_loopsplat,
  author={Liyuan Zhu and Yue Li and Erik Sandström and Shengyu Huang and Konrad Schindler and Iro Armeni},
  title     = {LoopSplat: Loop Closure by Registering 3D Gaussian Splats},
  booktitle = {International Conference on 3D Vision (3DV)},
  year      = {2025},
}

@inproceedings{sun2024mm3dgs,
  title={Mm3dgs slam: Multi-modal 3d gaussian splatting for slam using vision, depth, and inertial measurements},
  author={Sun, Lisong C and Bhatt, Neel P and Liu, Jonathan C and Fan, Zhiwen and Wang, Zhangyang and Humphreys, Todd E and Topcu, Ufuk},
  booktitle={2024 IEEE/RSJ International Conference on Intelligent Robots and Systems (IROS)},
  pages={10159--10166},
  year={2024},
  organization={IEEE}
}

@article{schubert2023visual,
  title={Visual place recognition: A tutorial [tutorial]},
  author={Schubert, Stefan and Neubert, Peer and Garg, Sourav and Milford, Michael and Fischer, Tobias},
  journal={IEEE Robotics \& Automation Magazine},
  volume={31},
  number={3},
  pages={139--153},
  year={2023},
  publisher={IEEE}
}

@article{campos2021orb,
  title={Orb-slam3: An accurate open-source library for visual, visual--inertial, and multimap slam},
  author={Campos, Carlos and Elvira, Richard and Rodr{\'\i}guez, Juan J G{\'o}mez and Montiel, Jos{\'e} MM and Tard{\'o}s, Juan D},
  journal={IEEE transactions on robotics},
  volume={37},
  number={6},
  pages={1874--1890},
  year={2021},
  publisher={IEEE}
}

@article{galvez2012bags,
  title={Bags of binary words for fast place recognition in image sequences},
  author={G{\'a}lvez-L{\'o}pez, Dorian and Tardos, Juan D},
  journal={IEEE Transactions on robotics},
  volume={28},
  number={5},
  pages={1188--1197},
  year={2012},
  publisher={IEEE}
}

@inproceedings{wang2025vggt,
  title={Vggt: Visual geometry grounded transformer},
  author={Wang, Jianyuan and Chen, Minghao and Karaev, Nikita and Vedaldi, Andrea and Rupprecht, Christian and Novotny, David},
  booktitle={Proceedings of the Computer Vision and Pattern Recognition Conference},
  pages={5294--5306},
  year={2025}
}

@article{maggio2025vggt,
  title={Vggt-slam: Dense rgb slam optimized on the sl (4) manifold},
  author={Maggio, Dominic and Lim, Hyungtae and Carlone, Luca},
  journal={arXiv preprint arXiv:2505.12549},
  year={2025}
}

@article{shi2024fast,
  title={Fast and accurate deep loop closing and relocalization for reliable LiDAR SLAM},
  author={Shi, Chenghao and Chen, Xieyuanli and Xiao, Junhao and Dai, Bin and Lu, Huimin},
  journal={IEEE Transactions on Robotics},
  volume={40},
  pages={2620--2640},
  year={2024},
  publisher={IEEE}
}

@article{cui2022bow3d,
  title={Bow3d: Bag of words for real-time loop closing in 3d lidar slam},
  author={Cui, Yunge and Chen, Xieyuanli and Zhang, Yinlong and Dong, Jiahua and Wu, Qingxiao and Zhu, Feng},
  journal={IEEE Robotics and Automation Letters},
  volume={8},
  number={5},
  pages={2828--2835},
  year={2022},
  publisher={IEEE}
}

@article{cattaneo2022lcdnet,
  title={Lcdnet: Deep loop closure detection and point cloud registration for lidar slam},
  author={Cattaneo, Daniele and Vaghi, Matteo and Valada, Abhinav},
  journal={IEEE Transactions on Robotics},
  volume={38},
  number={4},
  pages={2074--2093},
  year={2022},
  publisher={IEEE}
}

@inproceedings{sattler2019understanding,
  title={Understanding the limitations of cnn-based absolute camera pose regression},
  author={Sattler, Torsten and Zhou, Qunjie and Pollefeys, Marc and Leal-Taixe, Laura},
  booktitle={Proceedings of the IEEE/CVF conference on computer vision and pattern recognition},
  pages={3302--3312},
  year={2019}
}

@inproceedings{liu2024mvsgaussian,
  title={Mvsgaussian: Fast generalizable gaussian splatting reconstruction from multi-view stereo},
  author={Liu, Tianqi and Wang, Guangcong and Hu, Shoukang and Shen, Liao and Ye, Xinyi and Zang, Yuhang and Cao, Zhiguo and Li, Wei and Liu, Ziwei},
  booktitle={European Conference on Computer Vision},
  pages={37--53},
  year={2024},
  organization={Springer}
}

@inproceedings{takama2025sparse2dgs,
  title={Sparse2DGS: Sparse-View Surface Reconstruction Using 2D Gaussian Splatting with Dense Point Cloud},
  author={Takama, Natsuki and Ito, Shintaro and Ito, Koichi and Chen, Hwann-Tzong and Aoki, Takafumi},
  booktitle={2025 IEEE International Conference on Image Processing (ICIP)},
  pages={2844--2849},
  year={2025},
  organization={IEEE}
}

@inproceedings{chen2024mvsplat,
  title={Mvsplat: Efficient 3d gaussian splatting from sparse multi-view images},
  author={Chen, Yuedong and Xu, Haofei and Zheng, Chuanxia and Zhuang, Bohan and Pollefeys, Marc and Geiger, Andreas and Cham, Tat-Jen and Cai, Jianfei},
  booktitle={European conference on computer vision},
  pages={370--386},
  year={2024},
  organization={Springer}
}

@inproceedings{peng2025constrained,
  title={A Constrained Optimization Approach for Gaussian Splatting from Coarsely-posed Images and Noisy Lidar Point Clouds},
  author={Peng, Jizong and Tse, Tze Ho Elden and Xu, Kai and Gao, Wenchao and Yao, Angela},
  booktitle={Proceedings of the IEEE/CVF International Conference on Computer Vision},
  pages={2961--2970},
  year={2025}
}

@inproceedings{yan2024street,
  title={Street gaussians: Modeling dynamic urban scenes with gaussian splatting},
  author={Yan, Yunzhi and Lin, Haotong and Zhou, Chenxu and Wang, Weijie and Sun, Haiyang and Zhan, Kun and Lang, Xianpeng and Zhou, Xiaowei and Peng, Sida},
  booktitle={European Conference on Computer Vision},
  pages={156--173},
  year={2024},
  organization={Springer}
}

@inproceedings{zhou2024drivinggaussian,
  title={Drivinggaussian: Composite gaussian splatting for surrounding dynamic autonomous driving scenes},
  author={Zhou, Xiaoyu and Lin, Zhiwei and Shan, Xiaojun and Wang, Yongtao and Sun, Deqing and Yang, Ming-Hsuan},
  booktitle={Proceedings of the IEEE/CVF conference on computer vision and pattern recognition},
  pages={21634--21643},
  year={2024}
}

@inproceedings{detone2018superpoint,
  title={Superpoint: Self-supervised interest point detection and description},
  author={DeTone, Daniel and Malisiewicz, Tomasz and Rabinovich, Andrew},
  booktitle={Proceedings of the IEEE conference on computer vision and pattern recognition workshops},
  pages={224--236},
  year={2018}
}

\end{document}